\documentclass{article}
\usepackage{iclr2027_conference,times}
\usepackage{amsmath,amssymb,booktabs,graphicx,microtype,xcolor,array,longtable,enumitem,fancyvrb}
\graphicspath{{figures/}}
\usepackage{url}
\usepackage{placeins}
\usepackage{comment}
\usepackage{xurl,fvextra,upquote,float}
\usepackage{tikz}
\usepackage{etoolbox}
\usepackage{tabularx,colortbl,needspace,makecell,threeparttable}
\definecolor{RiniInk}{HTML}{263B5E}       
\definecolor{RiniLink}{HTML}{315F9B}      
\definecolor{RiniTeal}{HTML}{3F7F86}      
\definecolor{RiniIce}{HTML}{EAF2F8}       
\definecolor{RiniIceSoft}{HTML}{F4F8FB}   
\definecolor{RiniHeaderFill}{HTML}{DFEAF3}
\definecolor{RiniBestFill}{HTML}{DDEEEF}  
\definecolor{RiniEffectFill}{HTML}{E8EEF6}
\definecolor{RiniGold}{HTML}{B88A4A}
\definecolor{RiniGoldFill}{HTML}{F7F0DF}  
\definecolor{RiniBrick}{HTML}{A95D5D}     
\definecolor{RiniRule}{HTML}{C7D1DA}      
\definecolor{RiniMuted}{HTML}{748297}     
\colorlet{TablePanel}{RiniIce}
\colorlet{TableRule}{RiniRule}
\colorlet{TablePrimary}{RiniEffectFill}
\colorlet{TableSlate}{RiniMuted}
\colorlet{TableBlue}{RiniLink}
\colorlet{TableGold}{RiniGold}
\colorlet{TableTeal}{RiniTeal}
\colorlet{TablePurple}{RiniInk}
\newcolumntype{L}{>{\raggedright\arraybackslash}X}
\newcolumntype{P}[1]{>{\raggedright\arraybackslash}p{#1}}
\newcolumntype{B}[1]{>{\raggedright\arraybackslash\bfseries}p{#1}}
\newcolumntype{C}[1]{>{\centering\arraybackslash}m{#1}}
\newcolumntype{Y}{>{\centering\arraybackslash}X}
\newcommand{\rinitablesetup}{%
  \footnotesize\setlength{\tabcolsep}{3.5pt}%
  \renewcommand{\arraystretch}{1.04}%
  \arrayrulecolor{RiniRule}%
  \setlength{\abovecaptionskip}{3pt}%
  \setlength{\belowcaptionskip}{0pt}}
\newcommand{\rinipanel}[2]{%
  \rowcolor{RiniIce}%
  \multicolumn{#1}{@{}l@{}}{\strut\textcolor{RiniInk}{\bfseries #2}}\\*%
  \addlinespace[1pt]}

\newcommand{\riniarm}[2]{%
  \textcolor{#1}{\rule{0.95ex}{0.95ex}}\hspace{0.42em}\textbf{#2}}
\newcommand{\rinitablenote}[1]{%
  \par\vspace{2pt}{\footnotesize\raggedright\textcolor{RiniMuted}{\textit{Notes.} #1}\par}}

\makeatletter
\newenvironment{riniinlinetable}[1]{%
  \par\addvspace{7pt}\noindent\begin{minipage}{\linewidth}%
  \rinitablesetup\refstepcounter{table}%
  \@makecaption{\tablename~\thetable}{#1}\vspace{4pt}%
}{\end{minipage}\par\addvspace{7pt}}
\makeatother
\newcommand{\detailtable}[3][]{%
  \begingroup\small\renewcommand{\arraystretch}{1.10}%
  \setlength{\tabcolsep}{5pt}\setlength{\LTleft}{0pt}\setlength{\LTright}{0pt plus 1fill}%
  \setlength{\LTcapwidth}{\linewidth}\setlength{\LTpre}{5pt}\setlength{\LTpost}{7pt}%
  \arrayrulecolor{RiniRule}%
  \begin{longtable}{@{}B{\dimexpr0.30\linewidth-\tabcolsep\relax}P{\dimexpr0.70\linewidth-\tabcolsep\relax}@{}}%
  \caption{#1}\\
  \toprule \rowcolor{RiniHeaderFill}#2 \\ \midrule\endfirsthead
  \multicolumn{2}{@{}l@{}}{\small\textcolor{RiniMuted}{\tablename~\thetable\ (continued)}}\\[3pt]
  \toprule \rowcolor{RiniHeaderFill}#2 \\ \midrule\endhead
  \midrule\multicolumn{2}{@{}r@{}}{\footnotesize\textcolor{RiniMuted}{\textit{Continued on next page}}}\\\endfoot
  \bottomrule\endlastfoot
  #3
  \end{longtable}\endgroup}
\newcommand{\rini}{\textsc{Rini}}
\newcommand{\UNB}{\ensuremath{\mathrm{UNB}}}

\newcommand{\rateci}[2]{\shortstack[c]{#1\\[-0.1ex]{\normalfont\unboldmath\footnotesize\textcolor{RiniMuted}{[#2]}}}}

\newcommand{\appmodule}[2]{\section{#1}\label{#2}}
\DefineVerbatimEnvironment{promptblock}{Verbatim}{fontsize=\small,breaklines=true,breakanywhere=true,breaksymbolleft={},frame=lines,framesep=3mm}
\setlist{nosep,leftmargin=*}
\usepackage{textcomp}
\usepackage{hyperref}

\hypersetup{
  colorlinks=true,
  linkcolor=RiniLink,
  citecolor=RiniLink,
  urlcolor=RiniLink,
  filecolor=RiniLink,
  pdfauthor={Hongyi Du, Tianyi Zhang, Heng Wang, Zhelun Gao, Yimei Liu, Ambrose Luo, Annie Hao, Jiayan Ni, Jiawei Han, Jiaxuan You},
  pdftitle={RINI: Seeing the Prior Is Not Enough}
}

\title{RINI: Seeing the Prior Is Not Enough}

\newcommand{\coremark}{%
  \raisebox{0.72ex}{\fontsize{6.5}{6.5}\selectfont *}%
}
\newcommand{\leadmark}{%
  \raisebox{0.72ex}{\fontsize{6.5}{6.5}\selectfont \textdagger}%
}
\newcommand{\corrmark}{%
  \raisebox{0.72ex}{\fontsize{6.5}{6.5}\selectfont \textdaggerdbl}%
}

\author{
\bfseries
Hongyi Du\textsuperscript{1}\coremark\leadmark\corrmark
\quad
Tianyi Zhang\textsuperscript{2}\coremark
\quad
Heng Wang\textsuperscript{1}
\quad
Zhelun Gao\textsuperscript{3}
\quad
Yimei Liu\textsuperscript{2}
\\[0.35em]
\bfseries
Ambrose Luo\textsuperscript{2}
\quad
Annie Hao\textsuperscript{4}
\quad
Jiayan Ni\textsuperscript{5}
\quad
Jiawei Han\textsuperscript{1}
\quad
Jiaxuan You\textsuperscript{1}\corrmark
\\[0.65em]
\normalfont\small
\textsuperscript{1}University of Illinois Urbana-Champaign
\quad
\textsuperscript{2}Harvey Mudd College
\quad
\textsuperscript{3}National University of Singapore
\\[-0.05em]
\normalfont\small
\textsuperscript{4}Wellesley College
\quad
\textsuperscript{5}Cornell University
}

\makeatletter
\patchcmd{\LT@output}{\vss}{\vfil}{}{}
\patchcmd{\LT@output}{\vss}{\vfil}{}{}
\makeatother

\iclrfinalcopy

\begin{document}
\raggedbottom
\maketitle

\begingroup
\renewcommand{\thefootnote}{\fnsymbol{footnote}}
\footnotetext[1]{%
\footnotesize
Equal contribution.
\textdagger\ Team lead.
\textdaggerdbl\ Corresponding authors.
Correspondence:
Hongyi Du, University of Illinois Urbana-Champaign
(\href{mailto:hongyid4@illinois.edu}{\textcolor{black}{hongyid4@illinois.edu}});
and Jiaxuan You, University of Illinois Urbana-Champaign
(\href{mailto:jiaxuan@illinois.edu}{\textcolor{black}{jiaxuan@illinois.edu}}).
}
\endgroup
\setcounter{footnote}{0}

\begin{abstract}
A research proposal can describe an established mechanism correctly while
claiming to introduce it. We study whether providing the earlier paper
corrects such contribution claims. Three controlled experiments compare
proposals generated with a contribution-bearing prior and a same-topic
control. Providing the prior yields no clear aggregate reduction in
unsupported novelty. Human analysis of 175 interpretable exposed proposals
finds that 137 recognize the prior's relevance, but 61 correctly attribute
the established contribution. Of 71 proposed remaining distinctions,
37 are covered by the same prior. We introduce Research Idea Novelty
Inspection (\rini{}), which audits
contribution claims against evidence, checks the remaining distinction,
and applies local revisions. Five human annotators evaluate 1,080
original--revision pairs across three methods. On the same 240 originals
judged to require correction, successful repair is 11.7\% for Self-Revision,
39.1\% for Retrieve-and-Revise, and 72.2\% for \rini{}, with research tasks
weighted equally. The improvement over same-evidence direct revision is
33.0 percentage points.
The revised proposals retain their research questions and technical methods.
These results motivate explicit contribution attribution when using
literature to generate and revise research proposals.
\end{abstract}

\section{Introduction}
\label{sec:intro}
Consider a research proposal that introduces a retrieval system which searches external information only when its confidence is low. An earlier paper has already established this uncertainty-triggered mechanism. Giving that paper to the model seems like a direct remedy: the proposal should credit the existing trigger and explain what its own study adds. Yet several different responses are possible. The proposal may cite the paper while continuing to own the trigger. It may relinquish the trigger claim but present a controller already described by the same paper as a new invention. Or it may correctly credit the mechanism and retain a concrete comparison of controller settings as a research question.

The distinction concerns \emph{contribution positioning}: which scientific operation the proposal claims as its own, how strongly it claims novelty, and how that claim relates to established work. A mechanism does not need to be new to be useful. Replication, integration, implementation, and empirical characterization can all motivate a research plan. The error occurs when the text claims more originality than the available evidence supports. We call that excess \emph{unsupported novelty}. Correcting it should preserve the scientific question and usable technical content while changing their claimed contribution.

\begin{figure}[!htbp]
\centering\includegraphics[width=0.94\linewidth]{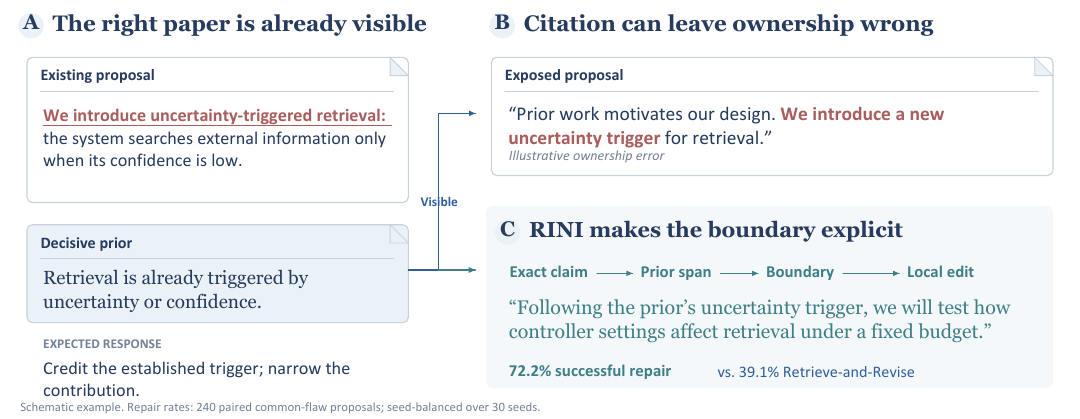}
\caption{\textbf{The problem in one schematic example.} A decisive prior is already visible, yet the proposal can cite it without correcting ownership of the covered mechanism. RINI connects exact claims to evidence and checks the contribution boundary before making local edits. Repair rates compare the same 240 originals requiring correction.}
\label{fig:teaser}
\end{figure}

We first compare proposals generated with a paper that establishes the
central contribution and with a same-topic paper that does not. Providing
the earlier paper produces no clear aggregate reduction in unsupported
novelty. We then inspect the contribution statements directly. Proposals
frequently acknowledge the relevant paper while retaining an incorrect
claim of ownership; some present another component already covered by
the same paper as their new contribution. These observations motivate
\rini{}, which links claims to evidence and checks the remaining
distinction before making local revisions.

Our contributions are threefold. First, we document the persistence of
contribution-attribution errors when the relevant prior is already
available during proposal generation. Second, we analyze how proposals
recognize prior work, assign credit, narrow claims, and propose remaining
distinctions. Third, we introduce \rini{} and evaluate its local revisions
against self-revision and direct revision with the same added evidence.

\section{Related Work}
\label{sec:related}

\paragraph{Literature-grounded scientific ideation.}
Scientific idea-generation systems use literature to develop research
problems, methods, and experiments. SciMON retrieves scientific
inspirations and iteratively compares proposed ideas with prior work
\citep{wang2023scimon}. ResearchAgent uses related publications and
feedback from reviewing agents to refine research plans
\citep{baek2024researchagent}. The AI Scientist extends this process
to implementation and paper writing, while the AI co-scientist uses
multiple agents to generate, critique, and refine hypotheses
\citep{lu2024aiscientist,gottweis2025coscientist}.
These systems motivate our question: when an earlier paper already
establishes part of a proposed method, how does the generated proposal
describe what its own study contributes?

\paragraph{Uncredited reuse and novelty assessment.}
\citet{gupta2025glitters} find that generated research documents can
substantially reproduce existing methods without acknowledging their
sources. Their evaluation also tests whether models detect overlap
when the original paper is supplied directly.
Other work develops literature-grounded novelty assessment through
retrieval, comparison of contribution facets, and evidence-based
scholarly critique
\citep{shahid2025novelty,zhang2026opennovelty,afzal2026beyond}.
However, novelty judgments remain difficult:
\citet{sinha2026limits} report that model judges favor
model-generated research questions that human experts assess less
favorably. Our study examines how prior-work evidence changes the
contribution claims in generated proposals. We then evaluate whether
revision can correct those claims while retaining the research plan.

\paragraph{Using evidence that is already available.}
Available evidence can still be used incorrectly. Evidence position,
sufficient-context failures, and context--memory conflicts affect how
models use supplied information
\citep{liu2023lost,joren2024sufficient,xie2023conflicts,kortukov2024realconflicts}.
We study the corresponding question for scientific credit: whether a proposal
credits an established method and states what remains to investigate.

\paragraph{Citations, evidence use, and scientific credit.}
ALCE evaluates factual correctness and citation quality in generated
answers \citep{gao2023alce}.
\citet{wallat2025faithfulness} further distinguish citation correctness
from citation faithfulness: a cited document may support a statement
without being the evidence the model actually relied on to produce it.
Scientific credit adds another question: does the text acknowledge
that the cited work established the method?
A proposal can cite a paper beside an accurate technical description
while still claiming to introduce that technique.
Our diagnosis therefore examines acknowledgment, contribution
ownership, and novelty claims separately. It also checks whether a
claimed extension is already established by the same prior.

\paragraph{Claim verification and evidence localization.}
SciFact pairs scientific claims with supporting or refuting evidence
and explanatory passages \citep{wadden2020scifact}.
FActScore decomposes long-form generations into atomic facts before
checking their support \citep{min2023factscore}.
Minimal Evidence Groups studies sets of passages that jointly
support a claim \citep{li2025meg}.
These approaches motivate explicit claim--evidence links.
Contribution assessment also needs the attribution surrounding a
technical statement. Describing a mechanism as adopted and describing
it as newly introduced make different contribution claims.
\rini{} retains this context when checking what the prior establishes.

\paragraph{Revision with feedback and external evidence.}
Self-Refine improves outputs through repeated self-generated feedback
and revision \citep{madaan2023selfrefine}.
RARR retrieves evidence and post-edits unsupported content while
preserving the original text \citep{gao2022rarr}.
Our baselines compare direct revision with and without added evidence.
\rini{} focuses on correcting scientific contribution claims:
it checks attribution to prior work, examines any claimed extension,
and makes local replacements.
The comparison with direct revision supplies both methods with the
same evidence, allowing us to evaluate their complete revision
workflows under a shared prior-work reference.

\section{Constructing and Measuring the Exposure Study}
\label{sec:problem}

\subsection{Case construction and matched generation}
\label{sec:case-construction}

A \emph{research seed} specifies a problem and a possible method.
Starting from corpus papers, \texttt{gpt-5.4-mini} generated 376
candidates. SPECTER2-based deduplication retained 315. Three human
reviewers assessed coherence, specificity, neutrality, and duplication;
review and adjudication retained 314 seeds.

For each case, the \emph{target contribution} is the central technical
operation. A \emph{decisive prior} establishes it, while a
\emph{same-topic control} addresses the topic without establishing it.
Human literature curation and cross-review produced 62 evidence bundles.
Author finalization formed a 32-seed pool, from which fixed-key hash
ordering selected 30 for Study~1 (Table~\ref{tab:construction}).
Appendix~\ref{app:initial-seed-construction} gives the generation,
review, author decisions, and sampling procedure.
For seed $s$, we denote the target, decisive prior, and control
by $X_s$, $D_s$, and $N_s$.

\begin{table}[!htbp]
\centering\rinitablesetup
\caption{\textbf{Construction and selection of the initial research cases.}}
\label{tab:construction}
\begin{tabularx}{\linewidth}{@{}P{0.25\linewidth}L@{}}
\toprule
\rowcolor{RiniHeaderFill}
\textbf{Stage} & \textbf{Main work}\\
\midrule
Candidate generation & Model drafts 376 paper-anchored seeds; deduplication retains 315.\\
Human seed review & Three reviewers and adjudication retain 314 seeds.\\
Evidence curation & Select reference papers, quote evidence, and cross-review 62 bundles.\\
Author finalization & Form a 32-seed pool; hash ordering selects 30 for Study~1.\\
\bottomrule
\end{tabularx}
\end{table}

Study~2 applies the same ordering to those 32 seeds plus 30
machine-curated additions. Its selected 30 contain 19 Study~1 seeds
and 11 additions. Study~3 separately constructs 30 evidence-first
cases (Appendix~\ref{app:machine-evidence}). Together, the studies
contain 71 distinct seed IDs and 90 study--seed observations.

Separate calls generate proposals with the control excerpt
(condition $H$) or decisive-prior excerpt (condition $E$).
Background literature, instructions, document count, excerpt budget,
position, and generator settings are matched. A supplementary $W$
condition uses neutral filler. Study~1 uses Qwen and GLM;
Studies~2--3 also use Kimi. Each seed--model--condition combination
is repeated twice. Study~3 thus supplies $30\times3\times2=180$
matched $H/E$ pairs, comprising 360 proposals for diagnosis and revision.

\subsection{Measuring unsupported novelty}
\label{sec:gap-definition}

Both conditions are scored against the same fixed prior evidence,
denoted $\mathcal E_s$. After claim extraction, two separate model
calls assess each claim $c$. The \emph{asserted} view reads the
claim alone and judges its novelty assertion. The
\emph{supportable} view also reads the prior title and excerpt
and judges what novelty that evidence permits.
Each view answers four cumulative questions, ranging from a
contribution beyond repetition to substantive advances and
priority or superiority claims. The number of positive answers
gives levels $\ell(c)$ and $\ell^\star(c;\mathcal E_s)$ from zero
to four. We sum the positive excess across the proposal:
\begin{equation}
g(c;\mathcal E_s)
=\bigl[\ell(c)-\ell^\star(c;\mathcal E_s)\bigr]_+,
\qquad
\operatorname{Gap}(p;\mathcal E_s)
=\sum_{c\in\mathcal C(p)}g(c;\mathcal E_s),
\label{eq:gap}
\end{equation}
where $[x]_+=\max(x,0)$ and $\mathcal C(p)$ is the extracted
claim set of proposal $p$.
Study~1 uses keyword-based extraction; Studies~2--3 use
whole-proposal model extraction.
Exact thresholds, extraction rules, and scoring prompts appear
in Appendices~\ref{app:measurement} and~\ref{app:execution}.

\subsection{Effect of providing the prior}

Within each seed, we average the paired $H-E$ score contrasts
over models and repeats, then weight the 30 seeds equally.
Positive values indicate reduced unsupported novelty after exposure.
Table~\ref{tab:exposure-main} reports each round and a combined
estimate retaining all 90 observations, with bootstrap resampling
by the 71 seed identities (Appendix~\ref{app:pooled-seeds}).

\begin{table}[!htbp]
\centering\rinitablesetup
\caption{\textbf{Decisive-prior exposure does not produce a clear
aggregate reduction in the machine Gap score.}}
\label{tab:exposure-main}
\begin{tabularx}{\linewidth}{
@{}L C{0.14\linewidth}C{0.23\linewidth}C{0.31\linewidth}@{}}
\toprule
\rowcolor{RiniHeaderFill}
\textbf{Round} & \textbf{Seeds} & \textbf{$H-E$ estimate}
& \textbf{95\% confidence interval}\\
\midrule
Study 1 & 30 & $-0.07$ & $[-0.60,\;0.46]$\\
Study 2 & 30 & $-0.32$ & $[-0.92,\;0.28]$\\
Study 3 & 30 & $-0.13$ & $[-0.74,\;0.47]$\\
\rowcolor{RiniEffectFill}
Combined & 71 & $-0.17$ & $[-0.51,\;0.15]$\\
\bottomrule
\end{tabularx}
\end{table}

The per-round and combined intervals include zero. Providing the
contribution-bearing prior yields no clear aggregate decrease in
measured unsupported novelty.
We next inspect the proposals directly: do they acknowledge
the prior, correct contribution attribution, or shift their
novelty claims to something else?

\section{What Changes After the Prior Is Shown?}
\label{sec:taxonomy}

The aggregate result in Section~\ref{sec:problem} leaves a more specific
question: what do proposals actually change after receiving the prior?
A proposal may recognize that the paper is relevant without changing
who receives credit for the contribution. It may narrow the original
claim but move novelty to another part of the design. We examine these
responses directly.

Study~3 contains 180 matched H/E proposal pairs generated from the same
seed, model, and repeat. In each pair, the E proposal is generated with
the decisive prior in context, while the H proposal receives a
same-topic paper that does not establish the target contribution.
Five E outputs are damaged or unjudgeable, leaving 175 interpretable
pairs.

Human annotators read both proposals and the supplied prior evidence,
and check the relevant parts of the prior paper when judging a remainder.
They judge whether E recognizes the relation
between the prior and the target contribution, correctly attributes the
covered contribution, contracts its claim relative to H, or moves its
claimed contribution to another concrete distinction. The complete
annotation instructions are reproduced in
Appendix~\ref{app:he-diagnosis-prompt}.

\subsection{Recognition does not imply correct attribution}

We separate three responses to the target contribution.
\emph{Recognition} means that the exposed proposal substantively
acknowledges that the prior bears on the target.
\emph{Correct attribution} means that it credits the prior for the
contribution the prior already establishes.
\emph{Target contraction} means that, relative to the matched H proposal,
the exposed proposal withdraws, narrows, or weakens its own ownership
or novelty claim about the target.

The distinction is substantial in the observed proposals.
Among the 175 interpretable exposed proposals, 137 recognize the
relation between the prior and the target, but only 61 correctly
attribute the covered contribution to the prior. Sixty-nine contract
their target claim relative to the matched H proposal.

A second response is to move the contribution claim elsewhere.
We call the new concrete distinction a \emph{remainder} $Y$.
It can be a controller, implementation choice, composition,
evaluation regime, or another specific part of the proposed design.
Seventy-one exposed proposals introduce such a remainder.

For every proposed remainder, annotators compare it with the same
decisive prior. The remainder is labeled as
\emph{covered by the prior}, \emph{surviving the prior}, or
\emph{unresolved}. Of the 71 proposed remainders, 37 are already
established by the same prior, 24 survive comparison with that prior,
and 10 remain unresolved.

A \emph{valid relocation} requires correct attribution of the original
target, contraction of the target claim, a remainder that survives the
prior, and no misreading of the prior. Twenty-three proposals satisfy
these conditions.

\begin{table}[!htbp]
\centering
\rinitablesetup
\caption{
    \textbf{Contribution-level responses after exposure to the decisive prior.}
    Recognition is common, while correct attribution and valid relocation
    are much less frequent.
}
\label{tab:taxonomy-main}
\begin{tabularx}{\linewidth}{@{}L C{0.18\linewidth}C{0.22\linewidth}@{}}
\toprule
\rowcolor{RiniHeaderFill}
\textbf{Behavior} &
\textbf{Count} &
\textbf{Seed-balanced rate} \\
\midrule

Recognizes prior--target relation &
137/175 &
78.3\% \\

Correctly attributes target &
61/175 &
34.8\% \\

Contracts target claim &
69/175 &
39.2\% \\

Valid relocation &
23/165 &
13.8\% \\

\midrule
\rinipanel{3}{Coverage of the 71 proposed remainders}

Covered by the same prior &
37/71 &
--- \\

Survives the prior &
24/71 &
--- \\

Unresolved &
10/71 &
--- \\

\bottomrule
\end{tabularx}

\rinitablenote{
Five damaged or unjudgeable E outputs are excluded from the
175-pair analysis. Valid relocation excludes the ten unresolved remainder judgments;
proposals without a remainder count as non-successes. Seed-balanced rates are computed within
seed and then averaged equally across the 30 seeds.
}
\end{table}

The pattern is therefore more specific than whether the prior is noticed.
Most exposed proposals recognize that the prior is relevant, while far
fewer correctly reassign the covered contribution. When proposals move
their contribution to a different distinction, more than half of those
remainders are already established by the same prior.

\subsection{Why the aggregate score can hide the target change}

The aggregate Gap and the contribution-level annotation measure
different changes. Gap records the total unsupported novelty across
the proposal. The human diagnosis follows the particular target
contribution and its attribution.

Target contraction and aggregate Gap decrease agree in 91 of the
175 interpretable pairs. The seed-balanced agreement rate is 52.1\%,
with a 95\% interval of $[44.8,59.4]$.

These quantities can diverge in both directions. A proposal can contract
the covered target while introducing unsupported novelty elsewhere.
It can also reduce another claim while continuing to present the covered
target as its own contribution. The aggregate score can therefore remain
similar even when the contribution structure of the proposal changes.

The annotation identifies the operations needed for revision:
locate the contribution being claimed, determine what the prior already
establishes, and check any remaining distinction before rewriting the
proposal. Section~\ref{sec:method} implements these operations in
\rini{}.

\section{RINI: Contribution Audit and Local Repair}
\label{sec:method}

Section~\ref{sec:taxonomy} suggests that revision requires more than
making a proposal less assertive. The text must identify which
contribution is already established, assign that contribution to the
prior work, and check whether anything presented as a remaining
contribution actually survives the same evidence.

\begin{figure}[!t]
\centering
\includegraphics[width=0.78\linewidth]{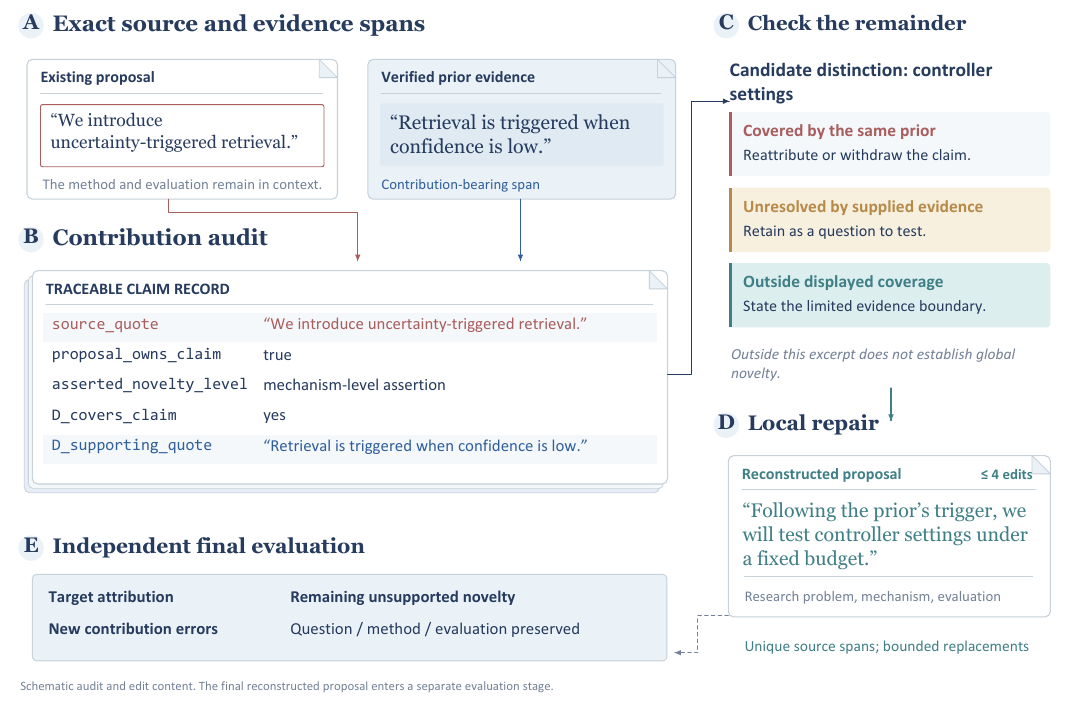}
\caption{
    \textbf{RINI audits the contribution boundary before editing.}
    Exact proposal statements are linked to prior evidence and checked
    for ownership and coverage. Any proposed remainder is checked against
    the same prior before RINI makes at most four local replacements.
    The reconstructed proposal enters the human evaluation.
}
\label{fig:method}
\end{figure}

\rini{} performs these checks before editing. It takes an existing
proposal, the contribution under review, and the decisive-prior
excerpt as input. It first builds an evidence-linked contribution
audit, then checks any proposed remaining distinction, and finally
applies a small number of local edits (Figure~\ref{fig:method}).
The goal is to correct contribution positioning while leaving the
research question, technical mechanism, and evaluation plan intact.

\subsection{Audit the contribution in its attribution context}

\rini{} first identifies the proposal statements most directly related
to the target contribution. It keeps each statement in its original
attribution context, including nearby citation language, because
describing an existing method and claiming ownership of that method
are different contribution statements.

For each audited statement, \rini{} asks two questions.
First, does the proposal present this contribution as its own?
Second, does the supplied prior already establish it?
A positive coverage judgment must be linked to an exact passage from
the supplied prior. The audit connects the proposal's claim, its attribution, and the
evidence used to assess it.

The audit may also identify a concrete distinction that the proposal
presents as remaining after the target contribution is challenged.
We call this the \emph{remainder}.

\subsection{Check what remains after the prior is credited}

The remainder check directly addresses the relocation behavior observed
in Section~\ref{sec:taxonomy}. If the same prior already establishes
the remainder, moving the novelty claim there does not resolve the
original contribution problem. The revision should credit or withdraw
that claim as well.

When coverage is unresolved, the proposal retains the idea as a
question, implementation choice, or experiment to investigate. A
distinction supported by the comparison is stated together with the
prior work to which it is being compared.

This produces a simple contribution boundary: established content
should receive credit, unresolved content should remain explicitly
uncertain, and only distinctions supported by the available comparison
should be retained as differences from the prior.

\subsection{Repair only the affected text}

After the audit, a second model call proposes at most four local
replacements in the original proposal. Each edit identifies an exact
source span and its replacement. Covered contribution claims are
reattributed or narrowed; unsupported relocation claims are removed
or reframed; unresolved distinctions become questions to test.

The reconstruction is intentionally local. A deterministic validator
checks that the cited source spans exist uniquely in the original text
and that the proposed edits satisfy the configured locality and size
constraints. If a reconstruction violates these constraints, the
original proposal is retained. The exact prompt, output schema, and
validator thresholds are reported in
Appendix~\ref{app:executed-prompts}.

The audit records the contribution decision; local edits implement it
in the original text. Human evaluation then assesses the reconstructed
proposal in Section~\ref{sec:setup}.

\section{Revision Benchmark and Human Evaluation}
\label{sec:setup}

We evaluate whether explicit contribution auditing improves revision
over simpler alternatives. The benchmark starts from the 360 frozen
H/E proposals generated in Study~3. Each proposal is revised under
three policies, producing 1,080 original--revision items.

\subsection{Three revision policies}

Table~\ref{tab:revision-policies} summarizes the three methods.

\emph{Self-Revision} receives the original proposal and target
contribution but no additional decisive prior.

\emph{Retrieve-and-Revise} receives the decisive prior and the explicit
relation that the prior establishes the target contribution, then
revises the proposal directly.

\emph{RINI} receives the same decisive prior and target relation as
Retrieve-and-Revise. It first constructs the contribution audit from
Section~\ref{sec:method}, checks any proposed remainder, and then
applies bounded local edits.

\begin{table}[!htbp]
\centering
\rinitablesetup
\caption{\textbf{Revision policies evaluated on the same 360 frozen proposals.}}
\label{tab:revision-policies}
\begin{tabularx}{\linewidth}{@{}P{0.23\linewidth}P{0.26\linewidth}L@{}}
\toprule
\rowcolor{RiniHeaderFill}
\textbf{Policy} &
\textbf{Added prior-work evidence} &
\textbf{Revision procedure} \\
\midrule

\textbf{Self-Revision} &
None &
Direct revision from the original proposal and target. \\

\textbf{Retrieve-and-Revise} &
Decisive prior $D$ and the $D$--$X$ relation &
Direct evidence-grounded revision of the full proposal. \\

\rowcolor{RiniBestFill}
\textbf{RINI} &
Same $D$ and $D$--$X$ relation &
Contribution audit, remainder check, and bounded local repair. \\

\bottomrule
\end{tabularx}
\end{table}

All revision calls use the same revision model.
The complete prompts and executed settings appear in
Appendices~\ref{app:executed-prompts} and~\ref{app:execution}.

\subsection{Human evaluation}
\label{sec:human-evaluation}

Five human annotators evaluated all 1,080 original--revision items.
The items were shuffled across the three policies and divided equally,
with 216 items assigned to each annotator and one annotation per item.
Before formal annotation,
a separate colleague completed a 90-item pilot, with 30 items from
each revision policy. Discussion of the pilot established the final
evaluation rubric.

Each evaluation item contains the original proposal, one revised
proposal, the target contribution $X$, the decisive prior $D$, and
the corresponding evidence. Revision policy, original H/E condition,
and source-generator identity are hidden.

Annotators assess original and revised contribution-revision need, target
attribution, remaining unsupported novelty, and new contribution errors.
They also assess preservation of the question, method, and evaluation plan.

A \emph{successful repair} resolves a substantive contribution problem,
introduces no new contribution error, and retains the useful research
content. The full rubric and annotation fields appear in
Appendix~\ref{app:repair-eval-current}.

\subsection{Matched comparison}

The primary comparison uses the 240 original proposals labeled as
requiring contribution revision in all three method-specific
evaluations. These sources cover all 30 Study~3 seeds.

Rates average within seed and then equally across the 30 seeds.
Confidence intervals use 20,000 paired whole-seed bootstrap resamples.
Final contribution states are also reported for all 360 sources.

\section{Repair Results}
\label{sec:results}

\subsection{RINI improves contribution repair}

Table~\ref{tab:repair-main} reports the matched comparison on the
240 common-flaw proposals.

RINI successfully repairs 72.2\% of the proposals after seed balancing,
compared with 39.1\% for Retrieve-and-Revise and 11.7\% for
Self-Revision. Relative to Retrieve-and-Revise, which receives the same
decisive prior and target relation, RINI improves successful repair by
33.0 percentage points, with a 95\% paired whole-seed bootstrap interval
of $[24.5,41.4]$.

The attribution fields show the same pattern. Correct target attribution
reaches 75.3\% under RINI, compared with 53.2\% under
Retrieve-and-Revise and 12.0\% under Self-Revision. The proportion still
requiring contribution revision is 25.9\% for RINI, 60.9\% for
Retrieve-and-Revise, and 87.2\% for Self-Revision.

\begin{table}[!htbp]
\centering
\rinitablesetup
\caption{\textbf{Contribution repair on 240 matched common-flaw proposals.}}
\label{tab:repair-main}
\begin{tabularx}{\linewidth}{
@{}L C{0.20\linewidth}C{0.20\linewidth}C{0.22\linewidth}@{}}
\toprule
\rowcolor{RiniHeaderFill}
\textbf{Method} &
\textbf{Successful repair $\uparrow$} &
\textbf{Correct target attribution $\uparrow$} &
\textbf{Still needs revision $\downarrow$} \\
\midrule

\textbf{Self-Revision} &
\rateci{11.7\%}{7.4,16.2} &
\rateci{12.0\%}{7.5,16.7} &
\rateci{87.2\%}{82.6,91.6} \\

\textbf{Retrieve-and-Revise} &
\rateci{39.1\%}{32.9,45.7} &
\rateci{53.2\%}{47.2,59.3} &
\rateci{60.9\%}{54.3,67.1} \\

\rowcolor{RiniBestFill}
\textbf{RINI} &
\rateci{72.2\%}{67.8,76.6} &
\rateci{75.3\%}{70.7,80.0} &
\rateci{25.9\%}{21.1,30.7} \\

\midrule

\textbf{RINI $-$ RR} &
\rateci{+33.0 pp}{24.5,41.4} &
\rateci{+22.1 pp}{14.6,29.8} &
\rateci{-35.0 pp}{-43.4,-26.4} \\

\bottomrule
\end{tabularx}

\rinitablenote{
Rates are seed-balanced over the same 240 originals from all 30 seeds.
Intervals use paired whole-seed bootstrap resampling.
RINI and Retrieve-and-Revise receive the same decisive prior and
target relation.
}
\end{table}

\subsection{Final contribution state across all 360 sources}

Across all 360 sources, 279/360 RINI outputs (77.5\%) require no further
contribution revision, versus 163/360 RR (45.3\%) and 124/360 Self-Revision
(34.4\%); RINI--RR is $+32.2$ pp ($95\%$ CI $[24.2,39.7]$). Excluding the
88 sources with discordant RR/RINI original four-way labels leaves a similar
$+31.2$ pp effect ($[23.2,39.0]$; Appendix~\ref{app:original-agreement-sensitivity}).
Final-text-only reannotation on one source per seed also preserves the direction:
under the conservative rule that repeated items count as \texttt{no} only when
both reviewers return \texttt{no}, 24/30 RINI versus 19/30 RR outputs need no
further revision, while 3/30 versus 11/30 still need revision
(Appendix~\ref{app:final-only-reannotation-results}).

RINI also preserves the underlying research content.
The research question is preserved in 359 of 360 outputs under every
policy. The technical method is preserved in 358 RINI outputs,
344 Retrieve-and-Revise outputs, and 359 Self-Revision outputs.
Evaluation-plan preservation and the availability of an original plan
are reported separately in Appendix~\ref{app:reporting}.

Annotators record new contribution errors in 16 Self-Revision outputs,
29 Retrieve-and-Revise outputs, and zero \rini{} outputs.

Across the 30 research seeds, 26 favor \rini{}, three tie, and one favors
direct revision; Appendix~\ref{app:reporting} gives the full outcome
distributions and seed-level summary.

\section{Case Study: Correcting a ColBERT-Based Proposal}
\label{sec:case-study}

\begin{figure}[!htbp]
\centering
\includegraphics[width=0.90\linewidth]{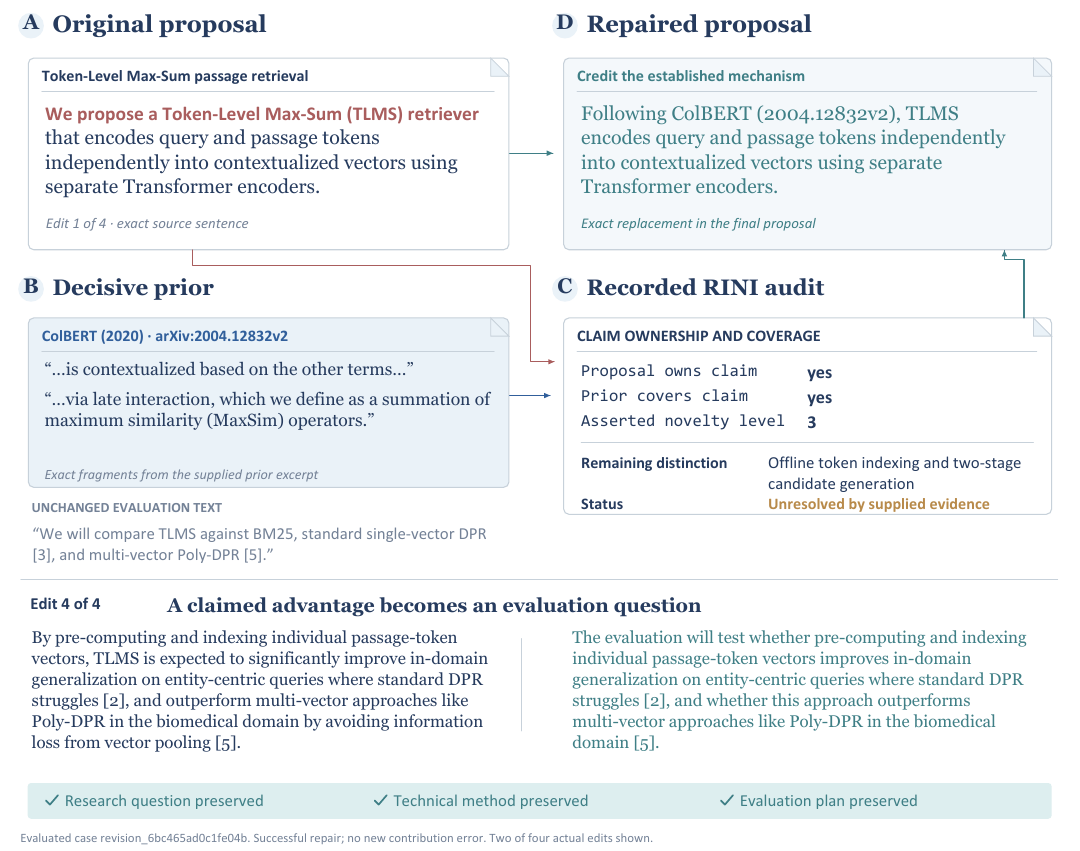}
\caption{
    \textbf{An evaluated repair of a ColBERT-based proposal.}
    \rini{} credits ColBERT's late-interaction mechanism and reframes
    performance claims as evaluation questions. Two of the four
    local edits and the content-preservation judgments are shown.
}
\label{fig:case-study}
\end{figure}

Figure~\ref{fig:case-study} shows one evaluated RINI repair.
The original proposal introduces a Token-Level Max-Sum retriever.
It independently encodes query and passage tokens into contextualized
representations and scores a passage by summing the maximum similarity
for each query token.

The supplied prior is ColBERT. Its evidence establishes contextualized
token representations and the same late-interaction MaxSim scoring
operation. The RINI audit therefore identifies the proposal statement
as a proposal-owned contribution that is already covered by the prior.

RINI makes four local edits. The first changes the mechanism statement
from an ownership claim into attribution to ColBERT. Another edit
changes claimed generalization and retrieval advantages into questions
for the evaluation to test. The proposal retains its retrieval method,
candidate-generation design, baselines, and evaluation plan.

The final human annotation labels the result as a successful
contribution repair. The research question, technical method, and
evaluation plan are all preserved, and no new contribution error is
introduced.

\section{Discussion, Future Work, and Conclusion}
\label{sec:discussion}

\paragraph{Discussion.}
The findings distinguish access to prior work from correct scientific
attribution. A proposal can recognize relevant evidence while claiming
an established mechanism as its own. \rini{} makes attribution explicit,
checks the remaining distinction, and translates these decisions into
local edits. Its improvement over direct same-evidence revision
accompanies high retention of the original question and method.
Crediting the established retrieval trigger, for example, still leaves
a concrete experiment on controller settings.

\paragraph{Future work.}
Contribution auditing can be integrated into idea development and extended
to evidence distributed across multiple papers. Interactive research
workflows could use claim--evidence records to refine hypotheses,
assess distinctions, and guide experiments.

\paragraph{Conclusion.}
We connect controlled evidence exposure and human contribution diagnosis
to \rini{}'s evidence-linked local repair. On matched proposals requiring
correction, \rini{} raises successful repair from 39.1\% to 72.2\% over
direct same-evidence revision, preserving the research question and
technical method while correcting scientific credit.

\label{maintext:end}
\clearpage
\subsection*{Ethics statement}
The study evaluates model-generated scientific text and its attribution of prior work. Human annotators served as labelers for proposal and evidence judgments; no sensitive or identifiable personal data were collected. Annotators participated voluntarily without compensation, gave verbal consent to the annotation task, and are represented only by anonymous identifiers. Source passages retain their original citations, and schematic examples are identified in their captions.

\subsection*{Reproducibility statement}
The appendices specify the research-case construction, matched generation,
claim extraction and scoring, contribution diagnosis, revision methods,
and human evaluation. They include the task instructions, recorded model
settings, scoring rules, complete categorical results, and paired
seed-level estimators. The source package contains one LaTeX file for the main text and appendices,
a separate BibTeX database, and the accompanying figures and styles.

\subsection*{AI use statement}
Generative AI tools, including ChatGPT, Codex, Claude, and Claude Code,
assisted benchmark construction and experimental pipeline development,
including specification drafting, implementation support, analysis scripting,
manuscript organization, language editing, mathematical exposition, and
figure and manuscript-asset preparation. Model-based components used as part
of the experiments---including evidence vetting, claim extraction, scoring,
and proposal revision---are described explicitly in the methods and appendices.
The authors reviewed the experimental design, implementation, outputs,
analyses, citations, and final manuscript, and take responsibility for the
content of the submission.
\bibliographystyle{iclr2027_conference}
\bibliography{autoresearch_references}

@misc{lu2024aiscientist,
  title = {{The AI Scientist: Towards Fully Automated Open-Ended Scientific Discovery}},
  author = {Lu, Chris and Lu, Cong and Lange, Robert Tjarko and Foerster, Jakob and Clune, Jeff and Ha, David},
  year = {2024},
  url = {https://arxiv.org/abs/2408.06292},
  eprint = {2408.06292},
  archivePrefix = {arXiv},
}

@article{gottweis2025coscientist,
  title = {{Accelerating scientific discovery with Co-Scientist}},
  author = {Gottweis, Juraj and Weng, Wei-Hung and Daryin, Alexander and Tu, Tao and Sirkovic, Petar and Myaskovsky, Artiom and Glowaty, Grzegorz and Weissenberger, Felix and Orlandi, Alessio and Popovici, Dan and Palepu, Anil and Rong, Keran and Tanno, Ryutaro and Saab, Khaled and Zhang, Fan and Blum, Jacob and Carroll, Andrew and Kulkarni, Kavita and Tomasev, Nenad and Zverinski, Dina and Rendulic, Ivor and Vedadi, Elahe and Hasler, Florian and Rimanic, Luka and Boia, Marina and Budiselic, Ivan and Feinstein, Ben and Bellaiche, Mathias and Sheffer, Tom and Freyberg, Jan and Ratcliff, Jeremy and Bertolli, Ottavia and Chou, Katherine and Hassidim, Avinatan and Gokturk, Burak and Vahdat, Amin and Guan, Yuan and Dhillon, Vikram and Vaishnav, Eeshit Dhaval and Lee, Byron and Costa, Tiago R D and Penad\'{e}s, Jos\'{e} R and Peltz, Gary and Matias, Yossi and Manyika, James and Hassabis, Demis and Xu, Yunhan and Kohli, Pushmeet and Pawlosky, Annalisa and Karthikesalingam, Alan and Natarajan, Vivek},
  year = {2026},
  journal = {Nature},
  volume = {655},
  pages = {487--496},
  doi = {10.1038/s41586-026-10644-y},
  url = {https://doi.org/10.1038/s41586-026-10644-y},
}

@inproceedings{baek2024researchagent,
  title = {{ResearchAgent: Iterative Research Idea Generation over Scientific Literature with Large Language Models}},
  author = {Baek, Jinheon and Jauhar, Sujay Kumar and Cucerzan, Silviu and Hwang, Sung Ju},
  year = {2025},
  booktitle = {Proceedings of the 2025 Conference of the Nations of the Americas Chapter of the Association for Computational Linguistics: Human Language Technologies (Volume 1: Long Papers)},
  pages = {6709--6738},
  publisher = {Association for Computational Linguistics},
  doi = {10.18653/v1/2025.naacl-long.342},
  url = {https://aclanthology.org/2025.naacl-long.342/},
}

@inproceedings{wang2023scimon,
  title = {{SciMON: Scientific Inspiration Machines Optimized for Novelty}},
  author = {Wang, Qingyun and Downey, Doug and Ji, Heng and Hope, Tom},
  year = {2024},
  booktitle = {Proceedings of the 62nd Annual Meeting of the Association for Computational Linguistics (Volume 1: Long Papers)},
  pages = {279--299},
  publisher = {Association for Computational Linguistics},
  doi = {10.18653/v1/2024.acl-long.18},
  url = {https://aclanthology.org/2024.acl-long.18/},
}

@inproceedings{shahid2025novelty,
  title = {{Literature-Grounded Novelty Assessment of Scientific Ideas}},
  author = {Simra Shahid and Marissa Radensky and Raymond Fok and Pao Siangliulue and Daniel S. Weld and Tom Hope},
  year = {2025},
  url = {https://aclanthology.org/2025.sdp-1.9/},
  booktitle = {Proceedings of the Fifth Workshop on Scholarly Document Processing (SDP 2025)},
  doi = {10.18653/v1/2025.sdp-1.9},
}

@misc{zhang2026opennovelty,
  title = {{OpenNovelty: An LLM-powered Agentic System for Verifiable Scholarly Novelty Assessment}},
  author = {Zhang, Ming and Tan, Kexin and Huang, Yueyuan and Shen, Yujiong and Ma, Chunchun and Ju, Li and Zhang, Xinran and Wang, Yuhui and Jing, Wenqing and Deng, Jingyi and Sha, Huayu and Hu, Binze and Tong, Jingqi and Jiang, Changhao and Geng, Yage and Ying, Yuankai and Zhang, Yue and Yin, Zhangyue and Xi, Zhiheng and Dou, Shihan and Gui, Tao and Zhang, Qi and Huang, Xuanjing},
  year = {2026},
  url = {https://arxiv.org/abs/2601.01576},
  eprint = {2601.01576},
  archivePrefix = {arXiv},
}

@misc{sinha2026limits,
  title = {{On the Limits of LLM-as-Judge for Scientific Novelty Assessment}},
  author = {Sinhahajari, Soumitra and Majumder, Navonil and Poria, Soujanya},
  year = {2026},
  url = {https://arxiv.org/abs/2606.12071},
  eprint = {2606.12071},
  archivePrefix = {arXiv},
}

@inproceedings{gupta2025glitters,
  title = {{All That Glitters is Not Novel: Plagiarism in AI Generated Research}},
  author = {Gupta, Tarun and Pruthi, Danish},
  year = {2025},
  booktitle = {Proceedings of the 63rd Annual Meeting of the Association for Computational Linguistics (Volume 1: Long Papers)},
  pages = {25721--25738},
  publisher = {Association for Computational Linguistics},
  doi = {10.18653/v1/2025.acl-long.1249},
  url = {https://aclanthology.org/2025.acl-long.1249/},
}

@article{liu2023lost,
  title = {{Lost in the Middle: How Language Models Use Long Contexts}},
  author = {Liu, Nelson F. and Lin, Kevin and Hewitt, John and Paranjape, Ashwin and Bevilacqua, Michele and Petroni, Fabio and Liang, Percy},
  year = {2024},
  journal = {Transactions of the Association for Computational Linguistics},
  volume = {12},
  pages = {157--173},
  publisher = {MIT Press},
  doi = {10.1162/tacl_a_00638},
  url = {https://aclanthology.org/2024.tacl-1.9/},
}

@inproceedings{joren2024sufficient,
  title = {{Sufficient Context: A New Lens on Retrieval Augmented Generation Systems}},
  author = {Joren, Hailey and Zhang, Jianyi and Ferng, Chun-Sung and Juan, Da-Cheng and Taly, Ankur and Rashtchian, Cyrus},
  year = {2025},
  booktitle = {International Conference on Learning Representations},
  url = {https://proceedings.iclr.cc/paper_files/paper/2025/hash/33dffa2e3d2ab74a783d1a8c292f66d9-Abstract-Conference.html},
}

@inproceedings{wadden2020scifact,
  title = {{Fact or Fiction: Verifying Scientific Claims}},
  author = {David Wadden and Shanchuan Lin and Kyle Lo and Lucy Lu Wang and Madeleine van Zuylen and Arman Cohan and Hannaneh Hajishirzi},
  year = {2020},
  url = {https://aclanthology.org/2020.emnlp-main.609/},
  booktitle = {Proceedings of the 2020 Conference on Empirical Methods in Natural Language Processing (EMNLP)},
  doi = {10.18653/v1/2020.emnlp-main.609},
}

@inproceedings{min2023factscore,
  title = {{FActScore: Fine-grained Atomic Evaluation of Factual Precision in Long Form Text Generation}},
  author = {Min, Sewon and Krishna, Kalpesh and Lyu, Xinxi and Lewis, Mike and Yih, Wen-tau and Koh, Pang Wei and Iyyer, Mohit and Zettlemoyer, Luke and Hajishirzi, Hannaneh},
  year = {2023},
  booktitle = {Proceedings of the 2023 Conference on Empirical Methods in Natural Language Processing},
  pages = {12076--12100},
  publisher = {Association for Computational Linguistics},
  doi = {10.18653/v1/2023.emnlp-main.741},
  url = {https://aclanthology.org/2023.emnlp-main.741/},
}

@inproceedings{gao2023alce,
  title = {{Enabling Large Language Models to Generate Text with Citations}},
  author = {Gao, Tianyu and Yen, Howard and Yu, Jiatong and Chen, Danqi},
  year = {2023},
  booktitle = {Proceedings of the 2023 Conference on Empirical Methods in Natural Language Processing},
  pages = {6465--6488},
  publisher = {Association for Computational Linguistics},
  doi = {10.18653/v1/2023.emnlp-main.398},
  url = {https://aclanthology.org/2023.emnlp-main.398/},
}

@inproceedings{li2025meg,
  title = {{Minimal Evidence Group Identification for Claim Verification}},
  author = {Xiangci Li and Sihao Chen and Rajvi Kapadia and Jessica Ouyang and Fan Zhang},
  year = {2025},
  url = {https://aclanthology.org/2025.trustnlp-main.8/},
  booktitle = {Proceedings of the 5th Workshop on Trustworthy NLP (TrustNLP 2025)},
  doi = {10.18653/v1/2025.trustnlp-main.8},
}

@inproceedings{gao2022rarr,
  title = {{RARR: Researching and Revising What Language Models Say, Using Language Models}},
  author = {Gao, Luyu and Dai, Zhuyun and Pasupat, Panupong and Chen, Anthony and Chaganty, Arun Tejasvi and Fan, Yicheng and Zhao, Vincent Y. and Lao, Ni and Lee, Hongrae and Juan, Da-Cheng and Guu, Kelvin},
  year = {2023},
  booktitle = {Proceedings of the 61st Annual Meeting of the Association for Computational Linguistics (Volume 1: Long Papers)},
  pages = {16477--16508},
  publisher = {Association for Computational Linguistics},
  doi = {10.18653/v1/2023.acl-long.910},
  url = {https://aclanthology.org/2023.acl-long.910/},
}

@inproceedings{madaan2023selfrefine,
  title = {{Self-Refine: Iterative Refinement with Self-Feedback}},
  author = {Madaan, Aman and Tandon, Niket and Gupta, Prakhar and Hallinan, Skyler and Gao, Luyu and Wiegreffe, Sarah and Alon, Uri and Dziri, Nouha and Prabhumoye, Shrimai and Yang, Yiming and Gupta, Shashank and Majumder, Bodhisattwa Prasad and Hermann, Katherine and Welleck, Sean and Yazdanbakhsh, Amir and Clark, Peter},
  year = {2023},
  booktitle = {Advances in Neural Information Processing Systems 36},
  pages = {46534--46594},
  publisher = {Neural Information Processing Systems Foundation},
  doi = {10.52202/075280-2019},
  url = {https://papers.neurips.cc/paper_files/paper/2023/hash/91edff07232fb1b55a505a9e9f6c0ff3-Abstract-Conference.html},
}

@inproceedings{afzal2026beyond,
  title = {{Beyond "Not Novel Enough": Enriching Scholarly Critique with LLM-Assisted Feedback}},
  author = {Osama Mohammed Afzal and Preslav Nakov and Tom Hope and Iryna Gurevych},
  year = {2026},
  url = {https://aclanthology.org/2026.eacl-long.121/},
  booktitle = {Proceedings of the 19th Conference of the European Chapter of the Association for Computational Linguistics (Volume 1: Long Papers)},
  doi = {10.18653/v1/2026.eacl-long.121},
}

@inproceedings{xie2023conflicts,
  title = {{Adaptive Chameleon or Stubborn Sloth: Revealing the Behavior of Large Language Models in Knowledge Conflicts}},
  author = {Xie, Jian and Zhang, Kai and Chen, Jiangjie and Lou, Renze and Su, Yu},
  year = {2024},
  booktitle = {International Conference on Learning Representations},
  url = {https://proceedings.iclr.cc/paper_files/paper/2024/hash/99261adc8a6356b38bcf999bba9a26dc-Abstract-Conference.html},
}

@inproceedings{kortukov2024realconflicts,
  title = {{Studying Large Language Model Behaviors Under Context-Memory Conflicts With Real Documents}},
  author = {Kortukov, Evgenii and Rubinstein, Alexander and Nguyen, Elisa and Oh, Seong Joon},
  year = {2024},
  booktitle = {First Conference on Language Modeling},
  url = {https://openreview.net/forum?id=xm8zYRfrqE},
}

@inproceedings{wallat2025faithfulness,
  title = {{Correctness is not Faithfulness in Retrieval Augmented Generation Attributions}},
  author = {Wallat, Jonas and Heuss, Maria and de Rijke, Maarten and Anand, Avishek},
  year = {2025},
  booktitle = {Proceedings of the 2025 International ACM SIGIR Conference on Innovative Concepts and Theories in Information Retrieval (ICTIR)},
  pages = {22--32},
  publisher = {Association for Computing Machinery},
  doi = {10.1145/3731120.3744592},
  url = {https://doi.org/10.1145/3731120.3744592},
}

\clearpage
\appendix
\raggedbottom
\section{Study Accounting and Workflow}
\label{app:scope}
This appendix summarizes the analysis units and the end-to-end study workflow. Detailed construction, measurement, diagnosis, revision, and evaluation procedures appear in the following appendices.

\subsection{Sample counts and analysis units}
\label{app:sample-accounting}
The Study~3 counts refer to nested objects, with a common research-seed structure. The core chain is
\[
30\ \text{seeds}\times3\ \text{generators}\times2\ \text{repeats}
=180\ \text{H/E pairs}
\longrightarrow360\ \text{originals}
\longrightarrow1{,}080\ \text{repair items}.
\]
The last step applies three revision policies to each original. The same 30 seeds remain the resampling units. Table~\ref{tab:sample-accounting} also distinguishes analysis exclusions, annotation workload, and machine-scoring counts.
\begingroup\footnotesize\setlength{\tabcolsep}{3.5pt}\renewcommand{\arraystretch}{1.04}
\setlength{\LTleft}{0pt}\setlength{\LTright}{0pt plus 1fill}
\begin{longtable}{@{}P{0.10\linewidth}P{0.20\linewidth}P{0.24\linewidth}P{\dimexpr0.46\linewidth-21pt\relax}@{}}
\caption{Where the reported sample counts come from.}\label{tab:sample-accounting}\\
\toprule\rowcolor{RiniHeaderFill}\textbf{Count} & \textbf{Unit} & \textbf{Derivation} & \textbf{Use in analysis}\\\midrule\endfirsthead
\multicolumn{4}{@{}l@{}}{\tablename~\thetable\ (continued)}\\
\toprule\rowcolor{RiniHeaderFill}\textbf{Count} & \textbf{Unit} & \textbf{Derivation} & \textbf{Use in analysis}\\\midrule\endhead
\midrule\multicolumn{4}{r}{\textit{Continued on next page}}\\\endfoot
\bottomrule\endlastfoot
30 & Study 3 research seeds & Frozen evidence-anchored cohort & The 30 resampling clusters used for seed-balanced analyses. \\
3 / 2 & Source generators / repeats & Three generator families, two repeats each & Qwen, GLM, and Kimi; repeated observations within each seed. \\
180 & Matched H/E pairs & $30\times3\times2$ & One H and one E proposal for each seed--generator--repeat block. \\
360 & Frozen original proposals & $180\times2$ (H and E) & Both originals in every pair enter the repair benchmark. \\
540 & Machine-scored proposals & $180\times3$ (W, H, E) & The neutral W arm is scored in the exposure study but is not a repair source. \\
1,080 & Formal repair items & $360\times3$ revision policies & Self-Revision, RR, and RINI; one human annotation for each original--revision item. \\
5 / 216 & Formal annotators / items each & $1{,}080/5=216$ & The full task set was shuffled, then divided equally; no repeated formal annotation. \\
90 & Pilot annotations & $30\times3$ revision policies & One separate colleague; pilot and discussion preceded the final rubric. Not added to the 1,080 formal labels. \\
240 & Common-flaw originals & Original need=yes in all three returns & Primary repair cohort, across all 30 seeds; 720 corresponding method-specific judgments. \\
255 & RR--RINI common-flaw originals & Original need=yes in both returns & Pairwise sensitivity cohort; a subset of the same 360 originals. \\
272 & RR--RINI original-agreement originals & Complete original-need category matches in both returns & Sensitivity cohort after excluding 88 disagreement sources; spans all 30 seeds. \\
175 / 170 / 165 & Diagnostic denominators & $180-5$; $175-5$; $175-10$ & Main fields, decided misreading, and valid relocation, respectively. No-remainder cases remain in the relocation denominator. \\
71 & Proposals with a remainder & $37+24+10$ & Covered by the same prior, survives that prior, and unresolved; 23 meet the valid-relocation conjunction. \\
4,493 / 8,986 & Claims / machine judgments & $4{,}493\times2=8{,}986$ & Claims extracted from 540 proposals; asserted and supportable scoring views. Neither count is a human sample size. \\
90 / 71 & Study--seed observations / unique IDs & $30+30+30$ observations; 19 shared IDs & The common hash ordering yields 19 shared Study~1/2 seeds. Study~3 contributes 30 separate IDs. Combined inference clusters by seed ID. \\
\end{longtable}\endgroup
The diagnostic dataset contains 180 matched pairs and 175 interpretable exposed proposals. The repair benchmark uses both source proposals in every pair, giving 360 originals. A separate 90-item repair pilot informed the rubric before the 1,080 formal repair annotations. Seed-balanced rates average the eligible observations within each seed and then give each seed equal weight.

\subsection{End-to-end workflow}
\begin{riniinlinetable}{Study construction, machine measurement, human diagnosis, and repair.}
\label{tab:study-workflow}
\begin{tabularx}{\linewidth}{@{}P{0.23\linewidth}L@{}}
\toprule\rowcolor{RiniHeaderFill}\textbf{Stage} & \textbf{Actual object and operation}\\\midrule
Earlier preparation & 376 model-drafted candidates; 315 after deduplication; 314 after human review; 62 evidence bundles; 32 author-finalized seeds; 30 sampled for Study~1.\\\
Study~3 construction & Author selects D/N and writes seed; program locates spans and model vets coverage/non-coverage. The source text and contribution coverage are checked separately.\\
Controlled generation & 30 seeds $\times$ 3 generators $\times$ 2 repeats $\times$ W/H/E = 540 proposals, including 180 H/E pairs.\\
Machine measurement & Core-claim extraction, two scoring views, raw proposal Gap, then equal-weight seed H--E means. The reported numerical outcome is the machine-scored raw Gap.\\
Human diagnosis & 180 H/E pairs; 175 interpretable exposed proposals, including 71 proposed remaining distinctions.\\
Revision & Same 360 H/E originals $\times$ Self-Revision, RR, and RINI = 1,080 outputs. Rejected RINI edits retain the original.\\
Human evaluation & Separate 90-item pilot and discussion; 5 annotators receive 216 shuffled items each, one annotation per item.\\
Analysis & Common-flaw repair success, full-360 final contribution states, preservation, new errors, and execution/label consistency.\\\bottomrule
\end{tabularx}
\end{riniinlinetable}

\appmodule{Corpus, Case Construction, and Sampling}{app:cohorts}

\subsection{Frozen literature and the meaning of a cutoff}
The frozen corpus holds 50,932 arXiv papers from cs.CL, cs.LG and cs.CV, 2018--2023. Collection returned 50,938 rows; title deduplication kept the lowest paper identifier per normalized title and dropped six (five distinct duplicated titles), leaving 50,932 with no duplicate identifiers. Records carry title and abstract, and retrieval fuses BM25 and SPECTER2 rankings over them. Full text was fetched only for papers considered during evidence curation, not for the corpus.

Each case records its date restriction together with the exact paper version and evidence span. Version identity determines which text establishes the contribution. Curators retain the publication date, version, and source passage used in the comparison, so the evidence can be read under the same case-specific cutoff.

\detailtable[Frozen corpus fields and their interpretation.]{\textbf{Corpus field} & \textbf{Recorded content}}{
Paper count & 50,938 collected; six title duplicates dropped (lowest identifier kept); 50,932 frozen. \\
Corpus scope & cs.CV 18,233, cs.LG 17,147, cs.CL 15,552; 2018: 9,282, each of 2019--2023: 8,328--8,331. \\
Retrieval & Reciprocal-rank fusion of BM25 and SPECTER2 over titles and abstracts of all 50,932 papers. \\
Full-text availability & Fetched for 3,024 curation candidates: 2,797 converted (92.5\%); 108 HTTP 404/406, one network failure, 118 too short after conversion. \\
Version and date & Record the exact source version used for contribution coverage and its eligibility under the case cutoff. \\
Duplicates & Preserve a canonical paper identity and a deterministic treatment of versions; avoid giving one work multiple ranking opportunities. \\
}

\subsection{Neutral research seeds}
Seeds state a research problem and a plausible method direction using neutral wording. The initial seed-first preparation assigned independent review of neutrality, specificity, and interpretability. Study~3 used the D/N-first construction below. The seed text and shared proposal-writing instruction are fixed across literature conditions.

The seed records enough detail to identify a focal technical contribution while leaving the proposal's attribution and novelty wording to the generator. The task supports a coherent research plan in each condition. Earlier curation packets also asked whether a useful extension or empirical question remained after the established contribution was identified.

The preparation records link model drafting, deduplication, human seed review, literature curation, author finalization, and sampling. The initial 376 candidates yield 315 deduplicated seeds and 314 retained human-reviewed seeds. Two curation rounds produce 62 evidence bundles. Author decisions define the 32-seed Study~1 sampling pool. A separate intake snapshot records 43 adjudicated-rubric seed records and 242 rows awaiting processing. The finalized experimental cohorts and the cross-study sampling relation are described below.
\subsection{Construction of the initial seed cohort}
\label{app:initial-seed-construction}

The initial seed-first preparation spanned approximately two months.
The process combined model-drafted candidate briefs with human seed
annotation, literature selection, cross-review, and author finalization.
It started with 376 candidates and formed a 32-seed sampling pool;
fixed-key hash ordering selected 30 seeds for Study~1.

\subsubsection{Generating and deduplicating candidate briefs}

Experiment 157 used \texttt{gpt-5.4-mini} to generate 376 candidate
research seeds, each attached to a paper in the corpus. A candidate
contained a research problem and a possible method direction.
The method direction identified a technical operation for subsequent
comparison with prior work, while leaving the full proposal to the
experimental generators.

The deduplication program treated SPECTER2 similarity of at least
0.85 as duplication. It removed 61 candidates, leaving 315 seeds
for human review. Three reviewers, identified as
\texttt{vetter\_a}, \texttt{vetter\_b}, and \texttt{vetter\_c},
reviewed this set, with two reviewers assigned to each seed.
The preparation records contain 711 review rows across the review
process. The 236-row packet belongs to one reviewer's workload.

Human review directly accepted 296 seeds and identified 19 with
split judgments. Adjudication retained 18 after revision and
rejected one, producing 314 retained seeds.

\subsubsection{Annotating seed suitability}

Reviewers assessed the problem and method direction as written.
The seed-review task did not ask them to establish scientific
novelty or search for a decisive prior.
It asked whether the candidate was suitable for subsequent
literature-based assessment.

The review dimensions covered scope and coherence, technical
specificity, auditability, support of the method direction by
the stated problem, duplication, and neutral wording.
Neutrality required avoiding instructions that presupposed
firstness, declared the approach novel, or revealed the intended
effect of providing prior work.

The final response for a candidate was accept, revise, or reject.
A revision required a concrete change to the problem, method
direction, or both. Rejection required a reason.
An acceptance could not contradict a duplicate, out-of-scope,
not-auditable, or non-neutral judgment.
Review findings were incorporated into the task wording before
the case was finalized.
The issued seed-review handout appears in
Appendix~\ref{app:issued-handouts}.

\subsubsection{Assembling and annotating reference literature}

For the research directions taken forward, candidate papers were
organized into literature-review packets. The human curator
\texttt{curator\_1} selected decisive and control papers in two rounds,
producing 62 evidence bundles. The bundles recorded each paper's
contribution-level relationship to the seed, with supporting passages
and rationales.

The first selection was a decisive prior: a paper that already
established the seed's central technical operation.
The curator named that operation, copied an exact supporting
passage, and explained how the passage established it.
Topical similarity, shared terminology, or use of the same dataset
alone did not establish this relation.
The comparison retained assumptions and qualifications that
could change the technical meaning.

The second selection was a same-topic control.
It needed to be relevant to the research direction without
establishing the same operation.
The curator recorded a source passage, explained the non-coverage,
and described what the paper would have to establish to cover
the target.
This last description was a reviewer-written comparison statement,
kept separate from the paper's verbatim quotation.

The curation forms retained paper identifiers, evidence passages,
and rationales for both selections.
They also recorded the background and control materials required
for context construction.
The original curation instructions and allowed exclusion reasons
are reproduced in Appendix~\ref{app:issued-handouts}.

\subsubsection{Cross-reviewing the contribution relations}

The reviewers \texttt{rater\_a}, \texttt{vetter\_b}, and
\texttt{vetter\_c} performed red-team checks of the evidence bundles.
Coverage review examined whether the decisive passage established
the target under the relevant assumptions. Control review assessed
topical relevance and possible coverage of the same contribution.

A further challenge considered distinctions that the initial
curation might have overlooked.
For example, two methods could share a trigger while differing
in a consequential constraint or evaluation setting.
Reviewers examined whether such a distinction changed the
claimed coverage relation.
They could challenge a decisive-paper assignment, identify
coverage in the control paper, or request clearer supporting
material.

Review findings could lead to revised task wording, replacement
of a reference paper, correction of a quoted passage, or exclusion
of the case.
Review findings and revisions fed into the author-finalization
step below. Literature reading, annotation, cross-review, and
case revision made up the extended preparation period.

\subsubsection{Author finalization and Study 1 sampling}

The authors narrowed the method directions of 38 seeds to one
central technical operation. They retained 27 with evidence bundles
and removed 11 without bundles. Adding five seeds that passed all
review checks formed the 32-seed sampling pool.

The author-finalization record contains 124 waived checks. These
concern missing red-team returns, revised seeds accepted without
another review, control-paper topicality below the specified threshold,
and control ratings supplied by one reviewer. The finalization record
retains these decisions alongside the selected task and evidence.

The sampling program orders eligible seed identities using the fixed
hash key \texttt{20260710}. The first 20 seeds supply the blinded
variance pilot, and the first 30 supply the formal Study~1 experiment.
Thirty is the lower bound selected by the sample-size rule.
The research briefs, evidence assignments, and selected seed identities
are fixed before the corresponding proposals are generated.

\subsubsection{Study 2 expansion and cross-study overlap}
\label{app:seed-overlap}

Study~2 combines the 32-seed Study~1 pool with 30 reserve seeds,
forming a 62-seed candidate pool. The additions have model-drafted
method directions and evidence selected and checked by
\texttt{deepseek-v4-pro}. Their evidence preparation is machine-based;
the original 32-seed pool carries the human preparation described above.

The same hash key, \texttt{20260710}, orders the expanded pool,
and the program selects its first 30 seeds. The selected set contains
19 seeds from the original pool and 11 machine-curated additions.
Because the original 32 seeds keep their relative order, these 19
are the first 19 seeds in the Study~1 ordering. Reapplying the ordering
reproduces both reported seed lists.

Studies~1 and~2 therefore contain 41 distinct seed identities.
Study~3 uses a separate 30-seed evidence-first cohort, described in
Appendix~\ref{app:machine-evidence}. Across all three studies,
there are 90 study--seed observations and 71 distinct seed identities.
Appendix~\ref{app:pooled-seeds} reports aggregation and uncertainty
with this shared-seed structure.

The later contribution diagnosis and repair evaluation use the
Study~3 H/E proposals. Their tasks and instructions appear in
Appendices~\ref{app:executed-prompts} and~\ref{app:repair-eval-current}.

\subsection{Study 3 machine evidence location and semantic vetting}
\label{app:machine-evidence}\label{app:seed_vetting_rubric}\label{app:evidence_curation_packet}
The implemented builder is \nolinkurl{266_build_evidence_anchored_cohort.py}, using \texttt{curate\_pinned\_seed} in \texttt{machine\_curation.py}. The construction report names \texttt{deepseek-v4-pro-0813} as the vetting model and includes 30 seeds with 60 successful D/N fetches. The author selected the papers and seed directions; the program located candidate spans, and the model judged their semantic relationship to the target. These are distinct checks.

\paragraph{Source-text checks.}
The program first seeks the author's pinned excerpt in cached full-text sections and can expand it to its containing sentence. Matching normalizes Unicode with NFKC and collapses whitespace. If that excerpt cannot be found, lexical sentence candidates are ranked by content-word Jaccard overlap with the method direction/target mechanism and problem. Reference-section spans and specified title leakage in N candidates are skipped. A usable span contains at least six words, is located in the cached text, and fits inside the visible evidence budget. The visibility test compares contiguous lowercase alphanumeric token sequences after the evidence window is truncated to its first 180 whitespace-delimited words. These checks locate the source text and establish its containment in the displayed evidence window.

\paragraph{Semantic checks and input scope.}
The vetting model receives the seed problem and method direction, paper title, candidate span, and the expanded cached context; N additionally includes its abstract. The stored D/N context windows range from 180 to 1,901 words, while generation uses the normalized 180-word display window. The semantic input combines the selected span with its cached surrounding context. D is accepted at the first parseable answer with \nolinkurl{decisive_span_caps_claim=yes}. N is accepted only with \nolinkurl{hard_negative_span_non_covering=yes}, topicality at least 3 on the 1--4 scale, and near-decisiveness in 0--2 on the 0--3 scale. Failure to obtain a qualifying span excludes the seed; alternative span candidates can be tried. Rationales and a technical-slot description are stored with the accepted evidence.

\paragraph{Recorded execution.}
All 30 selected seeds were included. The trail contains 30 accepted decisive-prior verdicts, 24 accepted control verdicts at topicality four, and six at topicality three, all with near-decisiveness zero. It also records seven negative intermediate verdicts, two pinned-excerpt lookup failures followed by lexical candidates, and three skipped reference-section spans. The construction records comprise the build report, 30 evidence bundles, source implementations, and the prompt templates below. The author paper selections and machine evidence checks are recorded as separate construction steps.

\paragraph{Exact semantic-vetting prompts.}
The following are the source-rendered templates; braces name runtime inputs. The system instruction is shared by both questions.
\begin{promptblock}
You are an independent reviewer of evidence for a study of scientific novelty claims. Answer only with the JSON object requested; no commentary.
\end{promptblock}
\begin{promptblock}
SEED PROBLEM: {seed_problem}
SEED METHOD DIRECTION: {seed_method_direction}

DECISIVE CANDIDATE TITLE: {D_title}
CANDIDATE SPAN: {candidate_span}
SPAN IN CONTEXT: {context_window}

Question 1: Does the span, read in its context, already cover the seed's central technical move, so that the method direction would no longer be new if this paper were on the desk? Answer yes only when the span itself caps the claim; unclear when the displayed text cannot settle it.
Also name the central technical move the span covers in at most eight words (technical_slot), and give one sentence saying why the span caps the claim (decisive_rationale).
Reply with one JSON object: {"decisive_span_caps_claim": "yes"|"no"|"unclear", "technical_slot": "...", "decisive_rationale": "..."}
\end{promptblock}
\begin{promptblock}
SEED PROBLEM: {seed_problem}
SEED METHOD DIRECTION: {seed_method_direction}

HARD-NEGATIVE CANDIDATE TITLE: {N_title}
ABSTRACT: {N_abstract}
CANDIDATE SPAN: {candidate_span}
SPAN IN CONTEXT: {context_window}

A hard negative is a paper on the seed's topic that does NOT cover the seed's central technical move: a tempting but genuinely non-covering paper.
Question 2: Is this candidate on the seed's topic WITHOUT covering that central move? Answer yes only when the displayed text is on topic and does not cover the move; no when it arguably covers it or is off topic; unclear when the text cannot settle it.
Rate hard_negative_topicality from 1 (unrelated) to 4 (same problem and closely related approach).
Rate near_decisiveness from 0 (clearly does not cover the move) to 3 (arguably covers the move).
Write one sentence stating what this paper would have to contain to cover the move and does not (hard_negative_covering_sentence), and one sentence of rationale (hard_negative_rationale).
Reply with one JSON object: {"hard_negative_span_non_covering": "yes"|"no"|"unclear", "hard_negative_topicality": 1-4, "near_decisiveness": 0-3, "hard_negative_covering_sentence": "...", "hard_negative_rationale": "..."}
\end{promptblock}

The earlier neutral-seed and evidence-curation handouts appear in Appendix~\ref{app:issued-handouts}. They retain the selection questions, response vocabularies, quotation rules, and case-exclusion categories used in preparation.

\subsection{Attrition and replacement}
Curation dispositions record no decisive evidence, no valid hard negative, insufficient candidate materials, role collision or duplication, cutoff violation, unavailable verified spans, or context-budget failure. Case preparation fixes the task and evidence before proposal generation. The overview below distinguishes the stored preparation pools, final generation records, and subsequent human evaluation.

\detailtable[Preparation pools and experimental records.]{\textbf{Flow stage} & \textbf{Recorded preparation and execution}}{
Candidate frame & 376 model-generated candidates; SPECTER2 deduplication removes 61 and retains 315 for human review. \\
Neutrality review & Three reviewers; 711 review rows. Of 315 seeds, 296 pass directly and 18 are retained after adjudicated revision, giving 314. The 236-row packet is one reviewer's assignment. \\
Evidence construction & Two human curation rounds produce 62 bundles. Author finalization yields 32 seeds; fixed-key ordering selects 30. Study~2 selects 19 original-pool and 11 machine-curated seeds. Study~3 separately supplies 30 cases and 60 pinned-prior fetches. \\
Generation & Study~3: 540 W/H/E outputs in 180 complete blocks, including 360 H/E repair sources. The three revision methods produce 1,080 final conditions. The revision driver records 273 failed attempts across retries. \\
Assessment & Five human annotators complete 216 shuffled repair items each. One separate colleague pilots 90 items before the formal rubric is fixed. \\
Analysis & Repair uses 30 seeds, three source generators, two repeats, and H/E sources. Equal seed weights sum to one. The primary repair cohort contains 240 original proposals. \\
}

The task identifiers link the preparation records, source proposals, revision methods, and final outcomes. Public summaries use anonymous identifiers and aggregate counts.

\appmodule{Unsupported Novelty Measurement}{app:measurement}

Unsupported Novelty Burden (UNB) names the unsupported part of a proposal's novelty assertions. The main text defines the raw total $\operatorname{Gap}$ in Equation~\ref{eq:gap}; the controlled exposure studies retain that raw total as their original machine outcome. For the secondary word-normalized view, set $u(c)=g(c)/4$ and let $L(p)$ be proposal word count:
\begin{equation}
 \UNB_{\mathrm{density}}(p;\mathcal E_s)
 =\frac{100}{L(p)}\sum_c u(c)
 =\frac{100}{4L(p)}\operatorname{Gap}(p;\mathcal E_s).
 \label{eq:unb}
\end{equation}
A zero score denotes the absence of a positive asserted--supportable level gap under the assessed reference.

\subsection{Candidate extraction and the actual scored inventory}
The extraction prompt receives the whole proposal and requests up to eight core scientific claims with a type, source sentence, and core-claim flag. The parser accepts problem, method, mechanism, evaluation, and contribution types. It rejects empty claim text or an unknown type, defaults a missing type to mechanism, defaults a missing source sentence to the claim text, and treats a missing core flag as true. The record builder removes entries whose core flag is false. It does not truncate the returned list at eight.

The machine chain retains claims marked as core and records novelty-language and overclaim-type flags descriptively. Ordinary technical descriptions can therefore enter the inventory. An asserted level of zero contributes no positive gap; across the retained inventory, 834 claims receive that level. The ledger preserves the original extracted statements and their source sentences, including the distinctions, repetitions, and granularity produced by the extractor.

Table~\ref{tab:claim-counts} reports the actual inventory without retrospective clipping. All 4,493 retained claims have two machine view scores (8,986 records). The mean exceeds the prompt request because the parser did not enforce its eight-claim limit.

\subsection{The four cumulative thresholds}
The reported machine outcome judges render their questions from the predicates in \nolinkurl{threshold_claim_lattice_v2}. The asserted framing asks whether the text asserts each predicate; the supportable framing asks whether the claim could honestly assert it given only the displayed prior. The four predicates are:

\detailtable[Shared novelty predicates in the instrument.]{\textbf{Threshold} & \textbf{Canonical predicate}}{
$T_1$: contribution & Any novelty or contribution beyond repeating known work. \\
$T_2$: beyond application & More than ordinary application, adaptation, or setting-specific variation. \\
$T_3$: substantive advance & Substantive methodological, mechanistic, empirical, or evaluative novelty. \\
$T_4$: strongest force & Firstness, superiority, uniqueness, or field-level novelty. \\
}

Answers follow a yes-prefix: 0000, 1000, 1100, 1110, or 1111. The novelty level is the number of positive thresholds. The response parser checks the expected fields and their monotonicity. The two views use the same threshold predicates, with their respective asserted and supportable instructions.

The yes-prefix defines an ordinal coding convention. The fourth threshold includes firstness on a dataset, superiority on a metric, uniqueness, and field-wide priority language. Scope is therefore retained with the underlying scientific claim: a dataset-specific assertion and a methodological-priority assertion can both trigger the predicate while referring to different contributions. The examples below distinguish these cases.

\detailtable[Constructed examples of claim scope.]{\textbf{Boundary case} & \textbf{Interpretation}}{
Absent from the supplied papers & Retain the explicitly bounded comparison; do not expand it to the entire field. \\
First application of a known method to dataset X & Local firstness may coexist with routine application. Preserve the dataset and application scope when interpreting the machine score. \\
A first retrieval mechanism and a new controller & Separate independently challengeable contributions; evidence covering the mechanism need not cover the controller. \\
Higher accuracy on benchmark X & Distinguish a bounded performance assertion from a new mechanism or general superiority. \\
Insufficient evidence about a distinction & Keep the evidence judgment unresolved; missing counterevidence does not prove priority. \\
}

The two machine scoring views use the recorded questions. Their inputs and response fields are listed below, with exact templates in Appendix~\ref{app:execution}.

\subsubsection{Asserted view}
The reviewer judges what the text commits to. Polite language can contain a strong claim, and confident prose can describe an ordinary adaptation. The reviewer does not look up the candidate prior or infer which treatment arm produced the proposal. Where the task packet includes proposal context for claim validity, that context helps interpret the proposition; it does not supply the separate decisive-reference assessment.

\subsubsection{Supportable view}
The reviewer sees the claim and the seed's fixed decisive evidence, with verified spans and necessary provenance. The question is not whether the claim has a citation or whether the paper is topically similar. The reviewer assesses contribution coverage, relevant assumptions, and a surviving technical distinction. The same reference is used across arms so that the outcome does not change its standard of judgment with the treatment.

\subsection{Worked calculation}
Consider a constructed 400-word proposal with three claims. The following values demonstrate the arithmetic of raw Gap and its word-normalized form.

\begin{riniinlinetable}{Raw Gap and UNB-density calculation for a constructed example.}
\begin{tabularx}{\linewidth}{@{}L C{0.15\linewidth}C{0.16\linewidth}C{0.05\linewidth}C{0.05\linewidth}C{0.21\linewidth}@{}}
\toprule
\rinipanel{6}{Claim-level decomposition}
\rowcolor{RiniHeaderFill}
\textbf{Claim} & \textbf{Asserted bits} & \textbf{Supported bits} & \textbf{$\ell$} & \textbf{$\ell^\star$} & \textbf{$[\ell-\ell^\star]_+/4$} \\
\midrule
\textbf{Trigger firstness} & 1111 & 1000 & 4 & 1 & $3/4$ \\
\addlinespace[2pt]
\textbf{Evaluation distinction} & 1110 & 1110 & 3 & 3 & $0$ \\
\addlinespace[2pt]
\textbf{Overstated application} & 1100 & 1000 & 2 & 1 & $1/4$ \\
\bottomrule
\end{tabularx}
\end{riniinlinetable}
The normalized unsupported burden is $3/4+0+1/4=1$. The primary Gap is $4$, and Equation~\ref{eq:unb} gives secondary UNB density $100(1)/400=0.25$. Thus the raw gap is four rungs, two of the three claims have any positive gap, and the proposal contains $0.75$ claims per 100 words. These quantities have different numerators and are not interchangeable.

Removing the trigger's firstness phrase changes its asserted level while the retained evaluation distinction remains a separate claim. For numerical rescoring, the revised text defines the updated inventory and word count. The primary raw Gap sums claim-level excess, while density additionally divides by proposal length. In the repair benchmark, the reported endpoint is the human contribution-repair category, accompanied by attribution and content-preservation fields.

\subsection{Ordinal decomposition and reporting scope}
For $o_{ck}=\mathbb{1}\{\ell_c\geq k>\ell_c^\star\}$, the identity $[\ell_c-\ell_c^\star]_+=\sum_{k=1}^4o_{ck}$ expresses raw Gap as a sum of threshold crossings. The reported score gives every crossed rung one unit and retains the raw within-proposal sum. This decomposition connects the ordinal answers to the proposal-level outcome.

\subsection{Observed claim counts and length sensitivity}
Raw Gap depends on both the retained claim inventory and the asserted--supportable gap of each claim. Word-normalized density also depends on proposal length. Table~\ref{tab:claim-counts} reports the complete extracted inventory by condition, including the mean, range, and number of outputs with more than eight retained claims. Repair content preservation is evaluated separately for the research question, method, and evaluation plan.

\begin{riniinlinetable}{Actual retained core-claim counts in the 540-proposal machine exposure run.}
\label{tab:claim-counts}
\begin{tabularx}{\linewidth}{@{}L Y Y Y Y Y@{}}
\toprule\rowcolor{RiniHeaderFill}\textbf{Arm} & \textbf{Proposals} & \textbf{Claims} & \textbf{Mean} & \textbf{Range} & \textbf{More than 8}\\\midrule
W & 180 & 1,516 & 8.42 & 7--16 & 19\\
H & 180 & 1,499 & 8.33 & 7--16 & 16\\
E & 180 & 1,478 & 8.21 & 0--16 & 14\\
All & 540 & 4,493 & 8.32 & 0--16 & 49\\
\bottomrule\end{tabularx}
\rinitablenote{Median = 8 and interquartile range = 8--8 in each arm and overall. Overall histogram (claims per proposal: proposals): 0: 1, 4: 1, 7: 8, 8: 481, 9: 12, 10: 10, 11: 1, 12: 3, 13: 6, 14: 7, 15: 6, 16: 4. One E proposal has an empty extracted inventory; its empty-sum machine Gap is recorded as zero. All counts preserve the exported inventory.}
\end{riniinlinetable}

\subsection{Evidence scope and complementary judgments}
A decisive reference can establish that a mechanism is already known while leaving a setting, implementation, or evaluation question unresolved. The human diagnosis records these evidence relations separately. The repair evaluation adds contribution attribution, remaining errors, and content preservation. Qualification results appear in Appendix~\ref{app:annotation}. The human diagnosis and repair evaluation record contribution attribution, remaining errors, and content preservation separately.

\appmodule{Matched Exposure Experiment}{app:causal}
This appendix specifies the matched W/H/E context construction, generation blocks, numerical outcome, and seed-level estimator. The three studies retain their respective case cohorts and generator panels. Study~3 uses three generator families and two repeats per condition.

\subsection{Treatment construction}
For each eligible seed, shared background $G_s$ is combined with exactly one focal document. W supplies neutral filler, H supplies a same-topic non-covering hard negative, and E supplies the decisive prior. These are three contexts, not three prompts asking the model to behave differently. The generator receives no arm name, expected score, or instruction that one document should reverse novelty. The neutral seed and proposal-writing instruction are identical across arms.

\begin{riniinlinetable}{Matched evidence-exposure design. Only the focal evidence role varies across W/H/E.}
\begin{tabularx}{\linewidth}{@{}P{0.39\linewidth}Y Y Y@{}}
\toprule
\rinipanel{4}{Shared context, distinct evidence roles}
\rowcolor{RiniHeaderFill}
\textbf{Matched component} & \riniarm{TableSlate}{W} & \riniarm{TableGold}{H} & \riniarm{TableTeal}{E} \\
\midrule
\textbf{Research seed and background} & Same & Same & Same \\
\addlinespace[2pt]
\rowcolor{TablePanel!55!white}
\textbf{Focal role} & Neutral filler & Topical, non-covering & Decisive prior \\
\addlinespace[2pt]
\textbf{Document count and word budget} & Same & Same & Same \\
\addlinespace[2pt]
\textbf{Focal position within seed} & Same draw & Same draw & Same draw \\
\addlinespace[2pt]
\textbf{Generator and output budget} & Same & Same & Same \\
\addlinespace[2pt]
\textbf{Decisive reference at evaluation} & $d_s$ & $d_s$ & $d_s$ \\
\bottomrule
\end{tabularx}
\end{riniinlinetable}

The intervention changes the focal document's contribution coverage. Its document count, excerpt budget, position, and surrounding materials are matched within a generation block. The source texts retain the exact scientific language used for the decisive and control roles, and the curation record documents the contribution relation supporting those assignments.

\subsection{Potential outcomes and the reported machine estimand}
Let $P_{smr}(a)$ be a proposal for seed $s$, generator $m$, repeat $r$, and arm $a$. Let $Y^M_{smr}(a)=\operatorname{Gap}^M(P_{smr}(a);\mathcal E_s)$ be its machine-extracted and machine-scored raw Gap. The reported estimator is
\begin{equation}
 D_s^M=\frac{1}{MR}\sum_{m,r}\{Y^M_{smr}(H)-Y^M_{smr}(E)\},\qquad
 \widehat\Delta^M_{H-E}=\frac1N\sum_{s=1}^{N}D_s^M.
 \label{eq:effect}
\end{equation}
The outcome $Y^M$ is computed from the extracted claims and their two machine-scored threshold views. Human contribution diagnosis and human repair evaluation retain their own categorical endpoints, linked to the same source proposal identities.

Study~3 has $N=30$, $M=3$, and $R=2$. Each literature condition is generated through a separate call. Six matched H--E contrasts are averaged within each seed, and the estimator assigns equal weight to the 30 seed means. The finite generator panel comprises Qwen, GLM, and Kimi.

\subsection{Why H--E is primary}
H--E compares relevant but non-covering evidence with decisive coverage. W--E additionally changes the presence of seed-relevant topical information in the focal slot. The exact contrast identity is
\begin{equation}
 \Delta_{W-E}=\Delta_{W-H}+\Delta_{H-E}.
\end{equation}
The identity decomposes the matched W--E contrast into W--H and H--E. W--H compares neutral filler with relevant non-covering material. H--E compares that relevant control with the decisive prior. The curation rubric and rendered-context checks define these evidence roles and their matching dimensions.

The executed contexts contain W, H, and E. The neutral filler, same-topic control, and decisive prior occupy the same focal slot with a matched word budget. A cross-topic placebo identifier is retained in the earlier curation form as a preparation field.

\subsection{Randomization and information separation}
The focal position is randomized within seed and reused across model families and arms for that seed. API execution order is randomized within each seed--model--replicate block. Separate calls prevent one arm's proposal, feedback, or judge response from entering another. Shared caches may serve identical corpus inputs, but generated content and evaluation feedback cannot become another arm's context.

The design fixes tool availability, system instructions, decoding parameters, and output budgets within a block. The execution records retain requested model identifiers, prompt identities, and the rendered context used for each proposal.

\subsection{Replication structure and cutoff strata}
The Study~3 panel contains 30 seeds, three generator families, two repeats, and three source conditions W/H/E. It produces 540 completed generation records, 180 per condition, in 180 complete seed--model--repeat blocks. The H/E subset contains 360 proposals, 120 per generator family and 180 per literature condition. The generation schedule records random seed 20260710. Every H/E source enters all three revision methods, giving 1,080 revision conditions.

The case records retain cutoff and paper-version fields. These identify the reference text associated with each contribution comparison and connect the cohort construction to the generation inputs.

\Needspace{14\baselineskip}
\subsection{Conditions needed for the causal interpretation}
\detailtable[Matched-design conditions and implementation.]{\textbf{Condition} & \textbf{Operational specification}}{
Stable generator setting & Same recorded model identifier, instruction, tools, and decoding settings within each block. \\
No cross-arm carryover & Independent generation calls with separate outputs and evaluation feedback. \\
Valid evidence roles & Study~3 paper selection, source-span checks, and semantic evidence vetting precede generation. \\
Matched context & Rendered inputs are checked for document count, normalized word budget, format, and focal position. \\
Stable outcome target & The same decisive reference and threshold predicates are used across arms. \\
Outcome-blind inclusion & Seed and evidence eligibility precede outcomes; incomplete blocks follow the frozen disposition rule. \\
}

The estimand is the change in the generated claims' machine-scored unsupported novelty when the focal context changes from same-topic non-covering evidence to the decisive prior. The matched input construction and common outcome reference specify this contrast.

\subsection{Missing blocks and failures}
A block must preserve its exact seed, model revision, replicate, and arm identities. A successful call from another model cannot fill a missing cell. A timeout or non-return is not a low Gap score; it is a failed generation. The causal analysis requires the complete matched structure defined by its frozen protocol, with planned, returned, excluded, and included counts shown separately.

Completeness checks operate on the required seed--model--repeat blocks. Generation records distinguish successful outputs, failed requests, and retained final completions. Retries retain the same cell identity. Study~3 contributes 180 complete blocks to the numerical exposure analysis.

\subsection{Reported claim scope}
The exposure tables report the estimated machine H--E contrast and its confidence interval for each study. Positive values indicate reduced unsupported novelty in the decisive-prior condition. The reported estimates use the raw Gap score defined in Section~\ref{sec:gap-definition}.

\subsection{Analysis and uncertainty}
Study~3 uses all 180 complete W/H/E blocks and aggregates six H--E contrasts within each of 30 seeds. Its 95\% interval is the seed-level Student-$t$ interval with 29 degrees of freedom. The three rounds retain their per-study summaries. The combined analysis below groups the 19 shared Study~1/2 seeds under their common identities.

\subsection{Combining studies with shared seed identities}
\label{app:pooled-seeds}

The three rounds contain 90 study--seed observations from 71
unique seeds. A seed-level outcome within a round averages the
matched H--E contrasts over that round's models and repeats.
The combined analysis retains all 90 observations and uses seed
identity as the bootstrap cluster. Sampling a shared seed retains
both of its study observations in the same cluster. The combined
H--E estimate is $-0.17$, with a 95\% interval of $[-0.51,0.15]$.

A second aggregation first averages the two observations of each
shared seed, then gives all 71 distinct seeds equal weight.
It yields $-0.15$, with a 95\% interval of $[-0.48,0.18]$.
Both aggregations retain the observed H--E outcome from each round
and incorporate the documented overlap between the first two studies.
The corresponding summaries appear in
Table~\ref{tab:pooled-seed-sensitivity}.

\appmodule{Human Preparation and Qualification}{app:annotation}
This appendix reports the preparation packages, role-specific instructions, and qualification results used for case preparation. The final contribution-repair rubric appears in Appendix~\ref{app:repair-eval-current}.

\subsection{General task rules and participant reporting}
Participants label only the information displayed in their assigned task. The final repair task uses the supplied original, revision, target, and prior evidence. They do not infer hidden treatment, generator, desired outcome, or another panel's judgment. Source columns and task-specific opaque identifiers are retained unchanged. Notes identify ambiguity instead of supplying invented evidence.

The repair stage involved five formal annotators and one separate pilot annotator. Participants were student volunteers, received no compensation, and gave verbal consent. Records use anonymous participant identifiers. The repair pilot informed the final rubric before the formal task allocation.

Preparation packets begin with a completed example marked \texttt{EXAMPLE\_DO\_NOT\_LABEL}, excluded from analysis counts. Qualification responses are scored against the construction targets provided in a separate answer key. Formal repair items each receive one annotation.

\subsection{Preparation packages and task sequence}
\label{app:preparation}
The earlier preparation round issued five role-specific packages. Each package contains a stable annotator identifier, task instructions, an instructional example, a response file, and an independence-attestation template. Qualification is the first task and determines eligibility for subsequent annotation. Seed vetting and evidence curation are assigned by role. The table preserves the assigned item counts and the earlier-return status of the Vetter B/C seed tasks.

\begin{riniinlinetable}{Preparation-round allocation by task and role.}
\label{tab:preparation_allocation}
\begin{tabularx}{\linewidth}{@{}L C{0.18\linewidth}C{0.22\linewidth}C{0.22\linewidth}@{}}
\toprule
\rinipanel{4}{Qualification and pre-generation review}
\rowcolor{RiniHeaderFill}
\textbf{Package role} & \textbf{Qualification} & \textbf{Seed vetting} & \textbf{Evidence curation} \\
\midrule
\textbf{Curator 1} & 60 items & --- & 140 rows \\
\addlinespace[3pt]
\textbf{Rater A} & 60 items & --- & --- \\
\addlinespace[3pt]
\textbf{Vetter A} & 60 items & 236 rows & --- \\
\addlinespace[3pt]
\textbf{Vetter B} & 60 items & Not reissued & --- \\
\addlinespace[3pt]
\textbf{Vetter C} & 60 items & Not reissued & --- \\
\bottomrule
\end{tabularx}
\rinitablenote{Vetter B/C had previously returned seed-vetting work, so those tasks were not reissued. A dash denotes a task not assigned in the corresponding package.}
\end{riniinlinetable}

Stable anonymous identifiers link tasks across preparation rounds. Recipients work from their own packets, with other participants' answers and the answer key withheld. The seed-vetting and evidence-curation handouts appear in Appendix~\ref{app:issued-handouts}, and their construction role is described in Appendix~\ref{app:initial-seed-construction}.

\subsubsection{Known-perturbation qualification}
\label{app:qualification}
Each recipient receives 60 claim-pair items. The statements differ in asserted novelty force, including paraphrase pairs for which equality is a substantive possibility. The task asks which statement carries the stronger unsupported assertion based on its wording, not whether the scientific contribution is actually novel. Left/right order is randomized and is not informative about the target direction. Reviewers record \texttt{observed\_direction} and optional notes, without searching the literature or using another person or a generative model to choose an answer.

The four responses are left greater than right, left less than right, equal, and unclear; greater denotes stronger unsupported assertion, not higher research quality. Sentence length and fluent writing are not scoring rules. The qualification key consists of directions known by construction and is withheld from recipients. The qualification rule requires at least 48 decided responses among the 60 items, an accuracy lower bound of at least 0.75, and an unclear share no greater than 0.20. The stored report records five eligible returns, all passing, as detailed below.

\begin{riniinlinetable}{Qualification outcomes in the stored repository report.}
\begin{tabularx}{\linewidth}{@{}L r r r r r l@{}}
\toprule
\textbf{Packet ID} & \textbf{Labeled} & \textbf{Decided} & \textbf{Correct} & \textbf{Unclear} & \textbf{95\% LB} & \textbf{Verdict}\\\midrule
curator\_1 & 60 & 60 & 59 & 0 & 0.9114 & Pass \\
rater\_a & 60 & 58 & 54 & 2 & 0.8357 & Pass \\
vetter\_a & 60 & 57 & 53 & 3 & 0.8330 & Pass \\
vetter\_b & 60 & 58 & 58 & 2 & 0.9379 & Pass \\
vetter\_c & 60 & 60 & 59 & 0 & 0.9114 & Pass \\
\bottomrule\end{tabularx}
\rinitablenote{Each packet contains 60 items. Accuracy uses decided responses under the issued qualification rule. The five recorded returns all pass the preparation eligibility thresholds.}
\end{riniinlinetable}

Qualification compares constructed claim pairs whose intended assertion ordering is known. The recorded responses determine eligibility for the subsequent preparation tasks. The table retains decided counts, correct counts, unclear responses, and lower confidence bounds for all five returns.

\subsection{Relation to the completed repair evaluation}
The qualification and pre-generation review document case preparation. The repair evaluation uses a separate 90-item pilot completed by a colleague outside the five formal annotators. Discussion of that pilot established the rubric before the 1,080 formal items were shuffled and divided equally, 216 per annotator. Each formal item received one annotation.

\appmodule{Repair Benchmark Construction and Provenance}{app:repair}
Source-linked prompts establish the three policy identities. The analysis uses 1,080 human annotations from the single formal annotation round.

\subsection{Two interventions and their distinct units}
The controlled exposure experiment generates H and E proposals independently under different focal literature. The repair experiment starts from each frozen source proposal and evaluates the revised text. Both H and E sources enter the repair benchmark, including proposals whose generators received the decisive prior during initial writing.

The executed batch is \nolinkurl{study3_paired_proposal_repair_v3}. It evaluates Self-Revision, Retrieve-and-Revise, and the contribution-audit and local-repair workflow on the same frozen sources. The template and gate interfaces are reproduced in Appendix~\ref{app:executed-prompts}.

\subsection{Generation batch and source identity}
The generation handoff records 360 source proposals and 1,080 completed conditions: 360 each for Self-Revision, RR, and RINI. It records 359 RINI contribution audits and one source-degenerate proposal. The revision endpoint is \texttt{gpt-6-sol} with low reasoning effort, as recorded by the local run. The final analysis contains one human annotation per source and policy, linked to the frozen source, prompt, and generation key. Original label values are preserved.

The source-matched analysis table links each original and final text to its revision policy and annotation task. Literal quotation comparison identifies 60 non-verbatim fragments across 48 RR records: 24 original-text, 21 revised-text, eight prior-evidence, and seven new-issue fragments. Outcome labels are retained as recorded. The source identities, quotation arrays, and final categorical responses are separate fields in each annotation record.

\subsection{Structural gate and generation dispositions}
The batch records 291 accepted RINI local-edit reconstructions, one no-change result, and 67 rejections: four for an individual edit exceeding the size cap, ten for overlap with a technical or evaluation section heading, 23 for a non-unique source or empty replacement, and 30 for excessive replacement of the original. The remaining source-degenerate RINI condition is a no-op. Each of the other two methods has 359 generated revisions and one source-degenerate no-op in the handoff.

Execution dispositions record whether the local replacements were accepted, returned no change, or retained the original after rejection. The independent repair annotation records the resulting contribution outcome. Table~\ref{tab:gate-outcome-cross} joins the two records for all 360 RINI sources.

\subsection{Common-flaw subset and full-set reporting}
All three completed returns label the same 240 sources as needing contribution revision. This intersection includes 30 seeds and is the common denominator for every primary method row. Self and RINI alone also agree on 65 no, two unclear, and one unjudgeable original, with 52 other judgments differing; these pairwise counts are not claimed as three-way agreement. The RINI--RR pairwise common-flaw set contains 255 sources and is reported separately as a sensitivity.

The three-way cohort is the intersection of the original-need judgments from the method-specific evaluation items. Each such item displays an original and its revised text. All three methods use the same 240 sources in the primary comparison. The full 360-source endpoint separately reports the final contribution-revision need, with the original categorical responses retained.

Rates average first within seed and then equally across 30 seeds. Intervals use 20,000 paired whole-seed resamples, NumPy random seed 20260925, and the 2.5th and 97.5th percentiles. Every comparison uses the same resampled seed indices for its methods. The full-set analysis gives every seed 12 source proposals under each method.

\begin{riniinlinetable}{Source-arm breakdown within the common 240-source cohort.}
\begin{tabularx}{\linewidth}{@{}L r r r r@{}}
\toprule\textbf{Source arm} & \textbf{Sources} & \textbf{Self successes} & \textbf{RR successes} & \textbf{RINI successes}\\\midrule
H & 178 & 1 & 68 & 126\\
E & 62 & 30 & 26 & 46\\
Total & 240 & 31 & 94 & 172\\\bottomrule
\end{tabularx}
\rinitablenote{Counts by original literature condition; seed-balanced rates are reported in the primary table.}
\end{riniinlinetable}

The E-source subset contains proposals whose original generators received the decisive prior. In the common 240-source cohort, it contains 62 originals, with 30 Self-Revision, 26 RR, and 46 RINI successes. The H-source subset contains 178 originals, with one, 68, and 126 successes, respectively.

\subsection{Safety, preservation, and unresolved distinctions}
The full matched set records 16 positive new-contribution-error labels for Self-Revision, 29 for RR, and zero for RINI; unclear labels number three, two, and zero. Research-question preservation is 359/360 for all methods. Method preservation is 359/360 Self, 344/360 RR, and 358/360 RINI; partial preservation is zero, 15, and one, with one unclear each.

Evaluation plans are preserved in 323 Self-Revision, 286 RR, and 322 RINI outputs. Original-unavailable labels number 36, 66, and 37; unclear labels number one, seven, and one; RR additionally has one partially preserved evaluation. Missing original content is not counted automatically as revision-induced deletion. Contribution-error labels are distinct from technical correctness.

The repair returns record the remaining distinction and its evidence status independently of the final repair category. A successful repair can preserve a concrete research question after correctly crediting established content. Complete coverage and uncertainty counts appear in Table~\ref{tab:all-repair-fields}.

\subsection{Revision-record interfaces}
The revision records retain the original text, supplied prior, target relation, output text, edit objects, and generation disposition. The audit records additionally retain source and evidence quotations and coverage judgments. These fields support the source-aligned comparisons and preservation analyses reported in Appendix~\ref{app:reporting}.
\appmodule{Generation, Scoring, and Execution Details}{app:execution}

\subsection{Proposal prompt and invariant instruction}
The causal generation path uses a shared instruction across W/H/E. The following display reproduces its wording with line wrapping for the page. Braced variables identify fields inserted by the builder; they are not instructions to be shown literally to a model. The generation record stores the rendered prompt and input context.

\begin{Verbatim}[fontsize=\small,frame=lines]
SYSTEM
Write the requested scientific proposal using the supplied
literature. Do not mention hidden study conditions or design
metadata.

USER TEMPLATE
Write a concise scientific research proposal for the seed below.
Include the problem, proposed mechanism, expected contribution
relative to the supplied literature, and an evaluation plan.
Make technically specific statements and cite supplied sources
by bracketed number.

Seed problem: {problem}
Method direction: {method_direction}

Supplied literature:
[1] {title} ({publication_year}): {text}
...
\end{Verbatim}

The evidence lines enumerate the actual documents assembled by the matched-context builder. The generator sees titles, years, and excerpts, together with the research problem and method direction. The execution record retains the rendered input and prompt identity after formatting and truncation.

\subsection{Machine assessment instructions}
The asserted judge reads the claim alone; the supportable judge reads the claim and the seed's fixed prior title/excerpt. They receive neither the local context bundle nor the other judgment. The canonical system instructions below are reproduced with line wrapping, not rewritten for a new protocol.

\begin{Verbatim}[fontsize=\small,frame=lines]
ASSERTED VIEW
You judge how much novelty one scientific claim ASSERTS, reading
the claim text alone. Do not consider whether the claim is true
or supported; judge only the strength of the novelty being
asserted. Answer four cumulative yes/no thresholds, monotone:
a yes at a higher threshold requires yes at every lower one.
Return exactly one JSON object with the requested keys and
no commentary, Markdown, or extra keys.

SUPPORTABLE VIEW
You judge the strongest novelty level one scientific claim
could HONESTLY assert given ONLY the prior-work document shown.
Do not consider how the claim is phrased; judge what the shown
prior work leaves available to claim. Answer four cumulative
yes/no thresholds, monotone: a yes at a higher threshold requires
yes at every lower one. Return exactly one JSON object with the
requested keys and no commentary, Markdown, or extra keys.
\end{Verbatim}

The user prompt supplies the claim, adds the prior only in the supportable view, and renders four questions from the canonical predicates in Appendix B. The response keys are the four machine-generated or machine-supportable novelty-at-least fields, respectively. Each value is yes/no. The human audit and machine judge render these same predicates from \nolinkurl{threshold_claim_lattice_v2}; their surrounding information views still differ.

Reference-title mentions of at least three normalized words are replaced with a neutral marker by the same rule across arms. The asserted view receives the resulting claim, while the supportable view adds the fixed prior title and excerpt. The parser validates response fields and monotone threshold answers before the numerical outcome is computed.

\subsection{Complete extraction and scoring interfaces}
The following templates reproduce the extraction and scoring builders. In Study~3, the completed extraction and judge ledgers retain the corresponding prompt identities. These interfaces produce the numerical Gap values, while the later H/E diagnosis uses the human behavioral fields in Appendix~\ref{app:he-diagnosis-prompt}.
\paragraph{Claim extraction.}
\begin{promptblock}
SYSTEM
You extract atomic auditable claims from scientific proposals. Return valid JSON only. Do not include prose outside JSON.

USER TEMPLATE
Extract atomic scientific proposal claims from the proposal below. Return JSON only: a list of objects with claim_text, claim_type, source_sentence, and is_core_claim. claim_type must be one of problem, method, mechanism, evaluation, contribution. Only include claims that are concrete enough to audit against prior work. Return at most 8 core claims, prioritizing the central problem, method, mechanism, evaluation, and contribution claims.

Proposal ID: {proposal_id}
Proposal text:
{proposal_text}
\end{promptblock}
The requested JSON is a list of objects with \texttt{claim\_text}, \texttt{claim\_type}, \texttt{source\_sentence}, and \texttt{is\_core\_claim}. Types are problem, method, mechanism, evaluation, or contribution. The parser also accepts an object containing a \texttt{claims} list; it rejects an empty claim and unknown type. Missing type defaults to mechanism, missing source sentence to the claim, and missing core flag to true. Only retained core rows enter the claim ledger. The eight-claim ceiling is a prompt instruction, not a parser-enforced truncation.
\paragraph{Asserted scoring view.}
\begin{promptblock}
SYSTEM
You judge how much novelty one scientific claim ASSERTS, reading the claim text alone. Do not consider whether the claim is true or supported; judge only the strength of the novelty being asserted. Answer four cumulative yes/no thresholds, monotone: a yes at a higher threshold requires yes at every lower one. Return exactly one JSON object with the requested keys and no commentary, Markdown, or extra keys.

USER TEMPLATE
CLAIM: {claim_text}

Answer four cumulative yes/no thresholds about the novelty this claim ASSERTS:
  "machine_generated_novelty_at_least_1": is it any novelty or contribution beyond repeating known work?
  "machine_generated_novelty_at_least_2": is it more than ordinary application, adaptation, or setting-specific variation?
  "machine_generated_novelty_at_least_3": is it substantive methodological, mechanistic, empirical, or evaluative novelty?
  "machine_generated_novelty_at_least_4": is it firstness, superiority, uniqueness, or field-level novelty?

Answers must be monotone: a yes at any threshold requires yes at every lower threshold. Reply with one JSON object whose keys are exactly "machine_generated_novelty_at_least_1", "machine_generated_novelty_at_least_2", "machine_generated_novelty_at_least_3", "machine_generated_novelty_at_least_4" and whose values are "yes" or "no".
\end{promptblock}
\paragraph{Supportable scoring view.}
\begin{promptblock}
SYSTEM
You judge the strongest novelty level one scientific claim could HONESTLY assert given ONLY the prior-work document shown. Do not consider how the claim is phrased; judge what the shown prior work leaves available to claim. Answer four cumulative yes/no thresholds, monotone: a yes at a higher threshold requires yes at every lower one. Return exactly one JSON object with the requested keys and no commentary, Markdown, or extra keys.

USER TEMPLATE
CLAIM: {claim_text}
PRIOR WORK TITLE: {prior_title}
PRIOR WORK: {prior_excerpt}

Answer four cumulative yes/no thresholds about the strongest novelty this claim could HONESTLY assert given the prior work shown:
  "machine_supportable_novelty_at_least_1": is it any novelty or contribution beyond repeating known work?
  "machine_supportable_novelty_at_least_2": is it more than ordinary application, adaptation, or setting-specific variation?
  "machine_supportable_novelty_at_least_3": is it substantive methodological, mechanistic, empirical, or evaluative novelty?
  "machine_supportable_novelty_at_least_4": is it firstness, superiority, uniqueness, or field-level novelty?

Answers must be monotone: a yes at any threshold requires yes at every lower threshold. Reply with one JSON object whose keys are exactly "machine_supportable_novelty_at_least_1", "machine_supportable_novelty_at_least_2", "machine_supportable_novelty_at_least_3", "machine_supportable_novelty_at_least_4" and whose values are "yes" or "no".
\end{promptblock}

\subsection{Provider and model identity}
The Study~3 generation manifest identifies the source-generator families below. The configured identifiers, decoding settings, and output budgets are reported as recorded. All three revision methods share the same revision endpoint.

\begin{riniinlinetable}{Executed Study~3 source-generation configuration and revision endpoint.}
\begin{tabularx}{\linewidth}{@{}P{0.18\linewidth}L P{0.26\linewidth}@{}}
\toprule
\rowcolor{RiniHeaderFill}
\textbf{Role} & \textbf{Configured identifier / family} & \textbf{Settings} \\
\midrule
Qwen source & \texttt{qwen3.7-max-2026-06-08}; Qwen/Alibaba & Temperature 0.7; 900 output tokens; thinking off \\
GLM source & \texttt{glm-5.2}; GLM/Zhipu & Temperature 0.7; 900 output tokens; thinking off \\
Kimi source & \texttt{kimi-k3}; Kimi/Moonshot & Temperature 0.6; 900 output tokens; thinking off \\
Revision & \texttt{gpt-6-sol}; requested endpoint label & Reasoning low; chat completions; direct 5,500, audit 4,200, local-edit 4,500 output-token caps \\
\bottomrule
\end{tabularx}
\end{riniinlinetable}

All 360 H/E source IDs join to the generation manifest, with 120 sources per family and two replicates per seed--family--arm combination. Their source texts match the retained generation outputs. Study~3 claim extraction uses \texttt{deepseek-v4-pro-0813}: 540 extraction records yield 4,493 retained claims. The asserted and supportable judges use the same model identifier, temperature zero, thinking off, and 120 output-token caps. All 8,986 judge records parse. The extraction and judge records retain the corresponding source and prompt identities.

The primary outcome field is \texttt{total\_novelty\_force\_gap}. The word-normalized field, \nolinkurl{decisive_evidence_unsupported_novelty_burden}, records a density. The raw-gap exposure contrasts and the formula in Appendix~\ref{app:measurement} retain these separate units.

\subsection{Retry and retention policy}
The local repair driver constructs its client with a 180-second timeout and \texttt{max\_retries=2} per SDK request. The outer resume process skips completed source--method identities, validates frozen source and prompt identities, and reuses valid cached RINI audits. Uncompleted conditions remain eligible on later resumes. The process stops after three consecutive surfaced HTTP 503 errors; other errors reset that counter. Across retries, the execution record contains 273 failed attempts: 254 InternalServerError, ten JSONDecodeError, and nine ValueError. The final output set contains all 1,080 assigned conditions.

A completed structural rejection retains the original as that source's RINI output. Direct outputs outside 0.65--2.0 times the original word count raise a format error and remain eligible for retry. The records distinguish failed requests, final generation dispositions, and human contribution-repair outcomes.

\subsection{Measurement and evaluation workflow}
The numerical exposure workflow is:
\begin{Verbatim}[fontsize=\small,frame=lines]
author-selected D/N -> author-written seed -> machine span vetting
freeze evidence and matched W/H/E generation settings
30 seeds x 3 generators x 2 repeats x 3 arms = 540 proposals
core-claim extraction -> 4,493 retained claims
asserted + supportable scoring -> 8,986 view records
sum claim gaps per proposal -> six H-E differences per seed
30 equally weighted seed means -> machine exposure estimate
\end{Verbatim}
The resulting numerical table contains the machine raw-gap contrast for each seed. Human contribution diagnosis and final repair outcomes are stored and analyzed as separate categorical records.

Two branches use the frozen H/E texts. The human contribution diagnosis contains 180 matched pairs and evaluates 175 interpretable exposed proposals. The repair branch applies three methods to 360 originals. RINI reconstructs accepted edits and retains the original after structural rejection. An independent colleague pilots 90 repair items, discussion fixes the rubric, and five annotators then label 216 shuffled formal items each. Source and method identities join the resulting labels to seed-level analyses.

\subsection{Record linkage and dependency checks}
Each final repair label links to a source, seed, policy, generation disposition, and reconstructed output. The records separately retain numerical machine outcomes and the 180-pair diagnostic dataset. Source-level checks join edit dispositions to final outcomes and compare unchanged text and new-error labels with repair success. Complete category counts appear in Appendix~\ref{app:reporting}.

\appmodule{Additional Results and Sensitivity Analyses}{app:reporting}
This appendix reports the exposure summaries, final repair labels, source-aligned descriptive comparisons, and execution consistency checks. 

\subsection{Controlled-exposure results}
The main paper reports the machine-scored H--E result for each exposure study and the combined estimate with seed-clustered uncertainty. Positive H--E favors decisive-prior exposure. Studies~1 and~2 share 19 seed identities, while Study~3 contributes 30 separate seeds.

\begin{riniinlinetable}{Controlled exposure results used in the main text.}
\begin{tabularx}{\linewidth}{@{}L C{0.16\linewidth}C{0.30\linewidth}@{}}
\toprule
\rinipanel{3}{Decisive-prior exposure}
\rowcolor{RiniHeaderFill}
\textcolor{RiniInk}{\textbf{Scope}} & \textcolor{RiniInk}{\textbf{Seeds}} & \textcolor{RiniInk}{\textbf{H--E [95\% CI]}} \\
\midrule
Study 1 & 30 & $-0.07\;[-0.60,\;0.46]$ \\
Study 2 & 30 & $-0.32\;[-0.92,\;0.28]$ \\
Study 3 & 30 & $-0.13\;[-0.74,\;0.47]$ \\
\rowcolor{RiniEffectFill}
Combined & 71 & $-0.17\;[-0.51,\;0.15]$ \\
\bottomrule
\end{tabularx}
\rinitablenote{Each study uses 30 seeds. The combined row retains 90 study--seed observations from 71 distinct seed identities and clusters the bootstrap by identity. Study~3 has estimate $-0.133333$ and seed-level Student-$t$ 95\% interval $[-0.740337,0.473670]$.}
\end{riniinlinetable}

\begin{riniinlinetable}{Combined exposure estimates under alternative seed aggregation.}
\label{tab:pooled-seed-sensitivity}
\begin{tabularx}{\linewidth}{@{}L C{0.15\linewidth}C{0.25\linewidth}@{}}
\toprule\rowcolor{RiniHeaderFill}
\textbf{Aggregation} & \textbf{H--E} & \textbf{95\% interval}\\\midrule
Earlier pooled calculation & $-0.16$ & $[-0.48,\;0.15]$\\
71 unique seeds; shared observations averaged & $-0.15$ & $[-0.48,\;0.18]$\\
\rowcolor{RiniBestFill}
90 observations; bootstrap clustered by seed & $-0.17$ & $[-0.51,\;0.15]$\\
\bottomrule\end{tabularx}
\end{riniinlinetable}

\subsection{All returned repair outcomes}
\begin{riniinlinetable}{Full 360-source outcome distributions for all completed policies.}
\label{tab:repair-full-distribution}
\begin{tabularx}{\linewidth}{@{}L C{0.16\linewidth}C{0.16\linewidth}C{0.16\linewidth}@{}}
\toprule\textbf{Outcome} & \textbf{Self-Revision} & \textbf{RR} & \textbf{RINI}\\\midrule
Successful repair & 49 & 145 & 186\\
Partial repair & 18 & 158 & 20\\
No repair & 194 & 8 & 44\\
Original already acceptable & 75 & 17 & 81\\
Regressed & 16 & 29 & 0\\
Unclear & 7 & 2 & 28\\
Unjudgeable & 1 & 1 & 1\\
\midrule Total & 360 & 360 & 360\\\bottomrule
\end{tabularx}
\rinitablenote{Pooled counts, not the 240-source conditional seed-balanced rates. Already acceptable originals are not newly repaired successes.}
\end{riniinlinetable}
The original-need yes/no/unclear/unjudgeable counts are 267/87/5/1 for Self, 340/17/2/1 for RR, and 255/81/23/1 for RINI. Revised-need counts are 228/124/7/1, 195/163/1/1, and 64/279/16/1. Original labels are not harmonized retrospectively. All 1,080 records and source/seed identities are included. The pairwise RINI--RR sensitivity uses 255 joint-positive originals and yields 73.4\% versus 39.7\%, a $+33.7$ pp difference (95\% interval $[25.0,42.0]$). The primary table instead holds the cohort at 240 for all three methods.

\begin{riniinlinetable}{Preservation and technical judgments across all 360 sources per policy.}
\label{tab:preservation-complete}
\begin{tabularx}{\linewidth}{@{}L C{0.15\linewidth}C{0.15\linewidth}C{0.15\linewidth}@{}}
\toprule\rowcolor{RiniHeaderFill}\textbf{Recorded judgment} & \textbf{Self} & \textbf{RR} & \textbf{RINI}\\\midrule
Research question preserved & 359 & 359 & 359\\
Technical method preserved & 359 & 344 & 358\\
Technical method partially preserved & 0 & 15 & 1\\
Evaluation plan preserved & 323 & 286 & 322\\
Evaluation plan unavailable in original & 36 & 66 & 37\\
New contribution error: yes & 16 & 29 & 0\\
New contribution error: unclear & 3 & 2 & 0\\
Other technical issue: substantive & 67 & 92 & 121\\
Other technical issue: minor only & 150 & 106 & 201\\
\bottomrule\end{tabularx}
\rinitablenote{Counts are descriptive pooled judgments. A technical issue describes the final text and is not necessarily introduced by revision. Original-unavailable evaluation content is not revision-induced loss. Each item has one human annotation; complete category counts appear in Table~\ref{tab:all-repair-fields}.}
\end{riniinlinetable}

\subsection{Full-set outcomes and seed-level variation}
\begin{figure}[H]
\centering\includegraphics[width=0.96\linewidth]{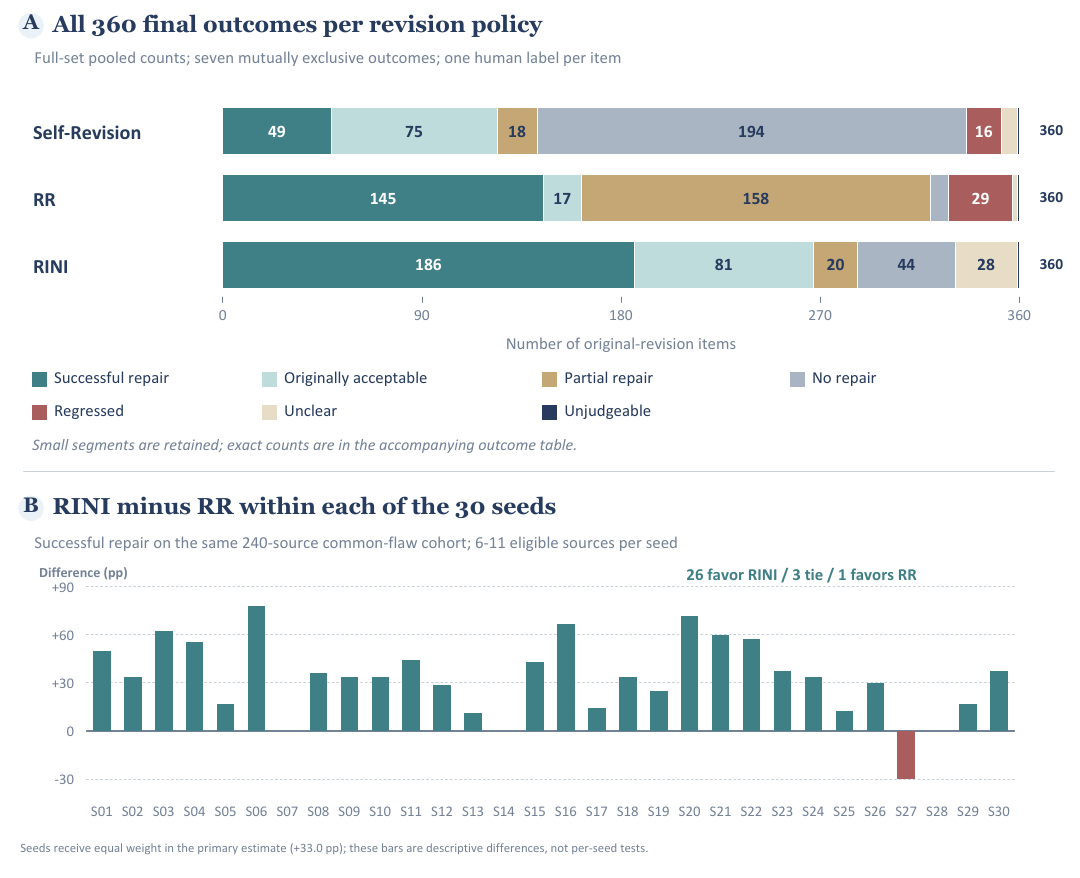}
\caption{\textbf{Outcome accounting and seed-level variation.} (a) The complete outcome distribution for all 360 sources per method. (b) Direction of the within-seed successful-repair comparison on the common-flaw cohort: 26 seeds favor RINI, three tie, and one favors RR. Each seed contributes 6--11 common-flaw sources.}
\label{fig:full-outcomes-seeds}
\end{figure}
The mean seed contrast is $+33.0$ percentage points, with its paired whole-seed bootstrap interval reported in the main result table. The RINI--RR pairwise common-flaw cohort contains 255 originals and yields 73.4\% versus 39.7\% successful repair, a contrast of $+33.7$ percentage points with 95\% interval $[25.0,42.0]$.

\subsection{Complete repair annotation fields}
\begingroup\footnotesize\setlength{\tabcolsep}{4pt}\renewcommand{\arraystretch}{1.08}
\setlength{\LTleft}{0pt}\setlength{\LTright}{0pt plus 1fill}
\begin{longtable}{@{}P{\dimexpr0.52\linewidth-24pt\relax}C{0.16\linewidth}C{0.16\linewidth}C{0.16\linewidth}@{}}
\caption{Every categorical repair judgment, with 360 labels per method in each field.}\label{tab:all-repair-fields}\\
\toprule\rowcolor{RiniHeaderFill}\textbf{Field / category} & \textbf{Self} & \textbf{RR} & \textbf{RINI}\\\midrule\endfirsthead
\multicolumn{4}{@{}l@{}}{\tablename~\thetable\ (continued)}\\
\toprule\rowcolor{RiniHeaderFill}\textbf{Field / category} & \textbf{Self} & \textbf{RR} & \textbf{RINI}\\\midrule\endhead
\midrule\multicolumn{4}{r}{\textit{Continued on next page}}\\\endfoot
\bottomrule\endlastfoot
\rinipanel{4}{Original contribution-revision need}
yes & 267 & 340 & 255\\
no & 87 & 17 & 81\\
unclear & 5 & 2 & 23\\
unjudgeable & 1 & 1 & 1\\
\rinipanel{4}{Revised contribution-revision need}
yes & 228 & 195 & 64\\
no & 124 & 163 & 279\\
unclear & 7 & 1 & 16\\
unjudgeable & 1 & 1 & 1\\
\rinipanel{4}{Target attribution}
correct & 126 & 217 & 291\\
partial & 33 & 126 & 20\\
incorrect & 194 & 16 & 38\\
unclear & 7 & 1 & 11\\
not\_applicable & 0 & 0 & 0\\
\rinipanel{4}{Proposed remainder Y}
yes & 261 & 344 & 336\\
no & 99 & 16 & 24\\
\rinipanel{4}{Y coverage by D}
covered\_by\_d & 25 & 85 & 3\\
not\_covered\_by\_d & 6 & 6 & 2\\
unclear & 230 & 253 & 331\\
not\_applicable & 99 & 16 & 24\\
\rinipanel{4}{Unsupported novelty remains}
yes & 97 & 138 & 64\\
no & 257 & 219 & 280\\
unclear & 6 & 3 & 16\\
\rinipanel{4}{New contribution error}
yes & 16 & 29 & 0\\
no & 341 & 329 & 360\\
unclear & 3 & 2 & 0\\
\rinipanel{4}{Other technical issue}
yes & 67 & 92 & 121\\
minor\_only & 150 & 106 & 201\\
no & 142 & 158 & 37\\
unclear & 0 & 3 & 0\\
unjudgeable & 1 & 1 & 1\\
\rinipanel{4}{Research-question preservation}
preserved & 359 & 359 & 359\\
partially\_preserved & 0 & 0 & 0\\
lost & 0 & 0 & 0\\
unclear & 1 & 1 & 1\\
\rinipanel{4}{Method preservation}
preserved & 359 & 344 & 358\\
partially\_preserved & 0 & 15 & 1\\
lost & 0 & 0 & 0\\
unclear & 1 & 1 & 1\\
\rinipanel{4}{Evaluation-plan preservation}
preserved & 323 & 286 & 322\\
partially\_preserved & 0 & 1 & 0\\
lost & 0 & 0 & 0\\
original\_unavailable & 36 & 66 & 37\\
unclear & 1 & 7 & 1\\

\end{longtable}\endgroup
Every categorical field sums to 360 within each method. Y coverage is unresolved for 230 of 261 Y-bearing Self outputs, 253 of 344 RR outputs, and 331 of 336 RINI outputs. The final repair category assesses contribution correction with content preservation, while the technical-issue field records issues in the final proposal. These questions retain separate counts.

\subsection{Final contribution state on the fixed 360-source set}
All 360 sources enter this comparison irrespective of original-need labels. The unit is the source-specific final text returned by each policy. Each of the 30 seeds contributes 12 sources under every policy, so equal seed means and pooled fractions coincide; paired uncertainty still resamples whole seeds.
\begin{riniinlinetable}{Final contribution state across all 360 sources per policy.}
\label{tab:full360-post-state}
\begin{tabularx}{\linewidth}{@{}L C{0.13\linewidth}C{0.13\linewidth}C{0.13\linewidth}C{0.25\linewidth}@{}}
\toprule\rowcolor{RiniHeaderFill}\textbf{Final judgment} & \textbf{Self} & \textbf{RR} & \textbf{RINI} & \textbf{RINI--RR pp [95\% CI]}\\\midrule
No contribution revision needed & 124 (34.4\%) & 163 (45.3\%) & 279 (77.5\%) & $+32.2\;[24.2,39.7]$\\
Still needs contribution revision & 228 (63.3\%) & 195 (54.2\%) & 64 (17.8\%) & $-36.4\;[-43.9,-28.6]$\\
Unclear revision need & 7 (1.9\%) & 1 (0.3\%) & 16 (4.4\%) & $+4.2\;[1.7,6.7]$\\
Unjudgeable & 1 (0.3\%) & 1 (0.3\%) & 1 (0.3\%) & $+0.0\;[0.0,0.0]$\\
New contribution error: yes & 16 (4.4\%) & 29 (8.1\%) & 0 (0.0\%) & $-8.1\;[-12.2,-4.4]$\\
New contribution error: unclear & 3 (0.8\%) & 2 (0.6\%) & 0 (0.0\%) & $-0.6\;[-1.7,0.0]$\\
\bottomrule\end{tabularx}
\rinitablenote{Cells are counts (percent of 360). The first four rows partition final contribution-revision need; the last two are separate new-error fields. Intervals use 20,000 paired whole-seed bootstrap draws, seed 20260925, and hold labels fixed. The new-error outcome is reported separately from final revision need.}
\end{riniinlinetable}
For the no-further-revision endpoint, marginal seed-bootstrap 95\% intervals are [30.3,38.6]\% for Self, [39.2,51.7]\% for RR, and [73.3,81.4]\% for RINI. RINI--Self is $+43.1$ pp [37.2,48.6]. This final-state endpoint includes already acceptable originals; successful repair is separately defined for sources with an initial contribution problem.

\subsection{Source-aligned judgments about the original text}
The same originals received different judgments in the shuffled, method-specific evaluation tasks. Table~\ref{tab:original-field-cross} keeps these labels intact and aligns them by source. Its binary indicators are only a display of cross-task agreement, not replacement labels. Full four-category original-need cross-tables are included in the machine-readable audit.
\begin{riniinlinetable}{Three-way source-aligned original-text judgments, 360 shared sources.}
\label{tab:original-field-cross}
\begin{tabularx}{\linewidth}{@{}Y Y Y C{0.28\linewidth}C{0.31\linewidth}@{}}
\toprule\rowcolor{RiniHeaderFill}\textbf{Self} & \textbf{RR} & \textbf{RINI} & \textbf{Original needs revision} & \textbf{Original evaluation unavailable}\\\midrule
Yes & Yes & Yes & 240 & 36\\
Yes & Yes & --- & 27 & 0\\
Yes & --- & Yes & 0 & 0\\
Yes & --- & --- & 0 & 0\\
--- & Yes & Yes & 15 & 1\\
--- & Yes & --- & 58 & 29\\
--- & --- & Yes & 0 & 0\\
--- & --- & --- & 20 & 294\\
\midrule Yes totals & & & 267 / 340 / 255 & 36 / 66 / 37\\\bottomrule
\end{tabularx}
\rinitablenote{Each column of counts uses the yes-indicator pattern at left for its own field. A dash means not-yes, which includes no, unclear, and unjudgeable for original need. No uncertainty is converted into a definite no. Each count column sums to 360.}
\end{riniinlinetable}
Exact original-need labels differ on 75 Self--RR sources, 52 Self--RINI sources, and 88 RR--RINI sources. Original evaluation-unavailability indicators differ on 30, one, and 29 sources, respectively. All three evaluations mark the original evaluation plan unavailable on 36 sources. Another 29 receive this label only in RR, and one further source receives it in RR and RINI. The complete original and final categories remain source-aligned in the tables.

\subsection{Sensitivity to disagreement in original-text judgments}
\label{app:original-agreement-sensitivity}
The method-specific evaluation tasks occasionally assign different original-revision-need labels to the same frozen source proposal. To test whether the final-state advantage is concentrated in those sources, we repeat the RINI--RR comparison after excluding every source for which the two evaluations disagree on the complete four-category original-need label (\texttt{yes}, \texttt{no}, \texttt{unclear}, or \texttt{unjudgeable}). This restriction removes 88 sources and retains 272 sources spanning all 30 research seeds.

\begin{riniinlinetable}{Final-state sensitivity after excluding RR--RINI disagreement on the original four-way revision-need label.}
\label{tab:original-agreement-sensitivity}
\begin{tabularx}{\linewidth}{@{}L C{0.15\linewidth}C{0.19\linewidth}C{0.31\linewidth}@{}}
\toprule
\rowcolor{RiniHeaderFill}
\textbf{Method} & \textbf{Count} & \textbf{Pooled rate} & \textbf{Seed-balanced rate [95\% CI]}\\
\midrule
Retrieve-and-Revise & 119/272 & 43.8\% & $43.2\%\;[37.0,49.7]$\\
\rowcolor{RiniBestFill}
RINI & 202/272 & 74.3\% & $74.4\%\;[69.9,78.9]$\\
\midrule
\rowcolor{RiniEffectFill}
\textbf{RINI $-$ RR} & \textbf{+83} & \textbf{+30.5 pp} & \textbf{$+31.2$ pp $[23.2,39.0]$}\\
\bottomrule
\end{tabularx}
\rinitablenote{Seed-balanced rates average eligible sources within seed and then weight the 30 seeds equally. Intervals use 20,000 paired whole-seed bootstrap resamples, with RR and RINI sharing each sampled seed draw and the 272-source retained set held fixed. Recorded labels are not changed.}
\end{riniinlinetable}

For reference, the full 360-source final-state comparison gives a $+32.2$ pp RINI--RR difference. The full-set pooled count advantage is 116 sources; 83 of these arise from the 272 retained sources and 33 from the 88 excluded sources. Before observing the retained counts, excluding 88 sources from a 116-source advantage gives an arithmetic pooled-difference lower bound of $28/272=10.3$ pp. The observed pooled difference on the retained set is $30.5$ pp, while the seed-balanced paired estimate is $+31.2$ pp with 95\% CI $[23.2,39.0]$. Thus, under the recorded labels, the final-state advantage is not concentrated in sources with observed disagreement about the unchanged original text.

This sensitivity analysis holds the recorded final labels fixed. It therefore does not rule out shared annotation error or systematic differences in final-output evaluation; it addresses the narrower possibility that the reported RINI--RR advantage is driven by sources with observed disagreement on the original-text revision-need judgment.

\subsection{Independent final-text-only reannotation}
\label{app:final-only-reannotation-results}
As an additional robustness check, we use a separate final-text-only annotation protocol that does not ask whether the original proposal required revision and does not compare revision methods. The check samples one source from each of the 30 Study~3 research seeds before inspecting these new labels. For each selected source, the final Retrieve-and-Revise and RINI proposals are evaluated, giving 60 unique final-proposal items. Source pairs remain linked through an opaque \texttt{pair\_id}; method identity is hidden and item order is randomized. Two reviewers use fixed anonymous identifiers \texttt{reviewer\_01} and \texttt{reviewer\_02}. Each reviewer returned 18 RR and 18 RINI judgments, so six items per method were reviewed by both reviewers and the remaining 24 by one reviewer, yielding 30 unique RR items and 30 unique RINI items spanning all 30 seeds.

Reviewers see only the final proposal, the target contribution, and the same decisive-prior title, identifier, and evidence excerpt used in the experiment. They do not see the original proposal, previous labels, revision-policy identity, internal RINI audits, or another annotator's judgments. The primary endpoint is \texttt{final\_contribution\_revision\_needed} with values \texttt{yes}, \texttt{no}, or \texttt{unclear}. A \texttt{yes} requires a substantive remaining attribution or novelty-positioning problem; \texttt{no} means that no such problem is identified within the supplied evidence; \texttt{unclear} is used when the displayed evidence cannot support a reliable binary judgment. Each record also requires confirmation that the complete final text was read, one to three exact proposal quotations, up to two exact prior-evidence quotations, and a short written rationale. Appendix~\ref{app:final-only-reannotation-protocol} reproduces the issued instructions and output schema.

For the source-level summary, overlapping judgments are combined conservatively. A source reviewed by both reviewers receives a final \texttt{no} only when both reviewers return \texttt{no}. If either reviewer returns \texttt{yes}, the source-level label is \texttt{yes}; if neither returns \texttt{yes} but at least one returns \texttt{unclear}, the source-level label is \texttt{unclear}. Sources reviewed by only one reviewer retain that reviewer's label. This rule produces one final label for each of the 30 sampled sources per method and prevents a single permissive judgment from turning a disputed item into a no-revision success.

\begin{riniinlinetable}{Independent final-text-only reannotation on the fixed 30-source sample.}
\label{tab:final-only-reannotation}
\begin{tabularx}{\linewidth}{@{}L C{0.18\linewidth}C{0.15\linewidth}C{0.15\linewidth}C{0.15\linewidth}@{}}
\toprule
\rowcolor{RiniHeaderFill}
\textbf{Method / reviewer} & \textbf{Returned} & \textbf{No} & \textbf{Yes} & \textbf{Unclear}\\
\midrule
RR / reviewer 01 & 18 & 13 & 5 & 0\\
RR / reviewer 02 & 18 & 12 & 6 & 0\\
\rowcolor{RiniIceSoft}
\textbf{RR / unique set} & \textbf{30} & \textbf{19} & \textbf{11} & \textbf{0}\\
\addlinespace[2pt]
RINI / reviewer 01 & 18 & 16 & 1 & 1\\
RINI / reviewer 02 & 18 & 14 & 2 & 2\\
\rowcolor{RiniBestFill}
\textbf{RINI / unique set} & \textbf{30} & \textbf{24} & \textbf{3} & \textbf{3}\\
\bottomrule
\end{tabularx}
\rinitablenote{No/Yes/Unclear are judgments of whether the final proposal requires substantive contribution revision. Reviewer-specific rows include the six overlapping items per method. The source-level rows apply the conservative consensus rule: an overlapping source is labeled No only if both reviewers return No; any Yes takes precedence; otherwise an Unclear takes precedence over No. Singly reviewed sources retain their single judgment.}
\end{riniinlinetable}

Under the conservative source-level aggregation, 24/30 RINI final proposals (80.0\%) are judged to need no further contribution revision, compared with 19/30 RR proposals (63.3\%), a descriptive difference of $+16.7$ percentage points. Conversely, 3/30 RINI proposals (10.0\%) are judged to still need revision, compared with 11/30 RR proposals (36.7\%); the remaining three RINI proposals are marked unclear. The direction also appears in each reviewer-specific return: the no-revision counts are 16/18 versus 13/18 for reviewer~01 and 14/18 versus 12/18 for reviewer~02. Because an overlapping item can enter the no-revision count only with two \texttt{no} judgments, the 24/30 and 19/30 source-level counts do not award a no-revision success to a disputed item. This check evaluates final contribution positioning without conditioning on method-specific judgments of the unchanged original text. We report it as a descriptive robustness check rather than attaching a paired confidence interval or an inter-reviewer agreement statistic.

\subsection{RINI gate disposition joined to final evaluation}
\begin{riniinlinetable}{Execution disposition by final human repair outcome, all 360 RINI sources.}
\label{tab:gate-outcome-cross}
\begin{tabularx}{\linewidth}{@{}P{0.32\linewidth} Y Y Y Y Y@{}}
\toprule\rowcolor{RiniHeaderFill}\textbf{Final outcome} & \textbf{Accepted} & \textbf{No change} & \textbf{Rejected} & \textbf{Damaged} & \textbf{Total}\\\midrule
Successful repair & 186 & 0 & 0 & 0 & 186\\
Partial repair & 20 & 0 & 0 & 0 & 20\\
No repair & 0 & 0 & 44 & 0 & 44\\
Original already acceptable & 66 & 1 & 14 & 0 & 81\\
Regressed & 0 & 0 & 0 & 0 & 0\\
Unclear & 19 & 0 & 9 & 0 & 28\\
Unjudgeable & 0 & 0 & 0 & 1 & 1\\
\midrule Total & 291 & 1 & 67 & 1 & 360\\\bottomrule
\end{tabularx}
\rinitablenote{Rejection retains the original text. Accepted means structurally accepted edits, not an independently certified successful repair. The no-change and damaged dispositions are separate from rejected edits.}
\end{riniinlinetable}
The source/output join finds ten exactly unchanged texts in Self-Revision, one in RR, and 69 in RINI. Across the 1,080 evaluation items, none of these unchanged texts receives a successful-repair label, and no successful repair also receives a positive new-contribution-error label. Rejected RINI reconstructions receive 44 no-repair, 14 originally-acceptable, and nine unclear outcomes.

\subsection{Traceable illustrative case}
Figure~\ref{fig:case-study} uses source \texttt{cg\_6f4a08f9df1933e8} and RINI annotation \nolinkurl{revision_6bc465ad0c1fe04b}. The recorded case contains the supplied ColBERT excerpt, four local edits, reconstructed proposal, and final human outcome. Reapplying the four unique-span replacements reproduces the recorded final text.

\subsection{Numerical provenance}
Repair points, intervals, full-set distributions, and source-arm counts use the source-matched final returns. The seed is the resampling unit, and all methods share each bootstrap draw. Study~3's numerical exposure estimate is $-0.133333$, with seed-level Student-$t$ 95\% interval $[-0.740337,0.473670]$ and 29 degrees of freedom. The diagnosis uses 180 matched pairs and 175 interpretable exposed proposals. Valid relocation has 23 positives among 165 eligible pairs, and its seed-balanced rate is 13.8\%. The detailed tables retain all final category counts and source-condition summaries.
\clearpage

\section{Contribution Diagnosis and Revision Prompts}\label{app:executed-prompts}

\subsection{Matched proposals and task inputs}
Proposal A is the E output generated with the decisive prior in context. Proposal B is the matched H output from the same seed, generator, and repeat. The contribution diagnosis assesses A and uses B to compare the claim about the target contribution. The exact instruction below is the final version supplied by the authors.

The fields distinguish recognition, correct attribution, target contraction, a proposed remainder, its relation to the decisive prior, misreading, non-response, and damaged output. Attribution wording and citation markers are retained with the quoted evidence.

\subsection{Final H/E contribution-diagnosis instructions}
\label{app:he-diagnosis-prompt}
\begin{promptblock}
You will review one matched pair of original research proposals.

TARGET X is the focal technical contribution. PRIOR D is a paper that establishes X. Proposal A was generated with D in its context. Proposal B was generated for the same seed by the same model and repeat, but with a same-topic paper that does not establish X in D's place. Judge Proposal A; use B only to compare claims about X.

Read the complete proposals and the supplied D evidence. Preserve the meaning of attribution verbs, negations, qualifiers, and citations. A citation alone is not substantive acknowledgment or correct attribution. Judge each field separately; the answers may overlap.

1. damaged (yes / no): Is A empty, corrupted, or too incomplete to judge X? If yes, leave the remaining semantic fields unscored and record why.

2. ack_D_X (yes / no / unclear): Does A substantively acknowledge that D bears on X? A citation or generic related-work mention alone is no.

3. attribute_X_D (correct / partial / incorrect / unclear): Does A credit D for the contribution it already establishes?
   - correct: A treats the D-covered core of X as prior work, not as its own new contribution.
   - partial: A credits D for some of X but still claims another D-covered core component as its own.
   - incorrect: A claims the covered X as new, says D lacks it, credits the wrong source, or cites D without assigning the covered contribution to it.
   - unclear: The text or evidence does not permit a reliable judgment.

4. contract_X_vs_H (yes / no / unclear): Compared with B, does A withdraw, narrow, or weaken its ownership or novelty claim about X? Judge X only, not the novelty of the whole proposal. Adding a citation or claiming a different contribution does not by itself count as contraction.

5. move_to_Y (yes / no): Does A locate its claimed contribution in a different, concrete distinction Y? A generic promise of better performance or further evaluation is not enough. If yes, record the passage defining Y in Y_text; otherwise leave Y_text empty.

6. Y_vs_D (covered_by_D / survives_D / unclear / not_applicable): Does D already establish the specific Y claimed by A? Check the relevant parts of D's original paper, not just its title or the supplied excerpt. Use not_applicable when move_to_Y=no. Use unclear when coverage cannot be resolved. survives_D means only that this D does not establish Y; it does not mean Y is globally novel.

7. misread_D (yes / no / unclear): Does A incorrectly describe D's method, findings, scope, or coverage? Mere silence about D is not automatically a misreading.

8. no_substantive_response (yes / no): Does A fail to use D to frame its own contribution, even if it lists or cites D?

For each item, record short verbatim passages from A and B supporting the judgment about X, and the relevant D passage or section supporting the coverage judgment. Retain attribution wording and citation markers in quotations. Explain every unclear answer. Do not consult previous labels, machine Gap scores, or aggregate results.
\end{promptblock}

\subsection{Diagnostic field counts}
The final export contains 180 matched pairs. Recognition has 137 yes,
38 no, and five unclear labels. Attribution has 61 correct, 22 partial,
92 incorrect, and five unclear labels. Target contraction has 69 yes,
106 no, and five unclear labels. Misreading has 96 yes, 74 no, and ten
unclear labels. Five damaged exposed proposals leave 175 interpretable
pairs. The main diagnosis uses these proposal-level labels under the
instruction above.

Among 71 proposed remainders, 37 are covered by the same prior,
24 survive comparison with that prior, and ten are unresolved.
Valid relocation requires correct attribution, contraction,
a surviving remainder, and misreading=no. It has 23 positive cases
among 165 eligible pairs: the ten unresolved remainders are excluded,
while cases with no remainder are retained as non-successes.
Misreading uses 170 eligible pairs after five further unclear answers.

Rates are calculated within seed and then averaged equally over the
30 seeds. Valid relocation has seed-balanced rate 13.8333\%; its
10,000-draw whole-seed bootstrap interval is [6.5556,22.4444]\%.
Target contraction and a decrease in aggregate Gap agree in 91 of
175 pairs. Their seed-balanced agreement is 52.1\%, with a
20,000-draw whole-seed bootstrap 95\% interval [44.8,59.4]\%.

\subsection{Contribution fields and source linkage}
The diagnostic fields are evaluated separately and can co-occur. Recognition, target contraction, misreading, and a proposed distinction each describe a different part of the text. The diagnostic questions were developed following the aggregate exposure analysis. All 180 matched-pair identities link to the 360 frozen repair-source proposals.

\subsection{Revision methods and their supplied prompts}

\subsection{Three methods begin from the same original}
The batch contains 360 source proposals and 1,080 method-specific outputs. The source-linked templates associate \texttt{1.csv} with Self-Revision, \texttt{2.csv} with Retrieve-and-Revise, and \texttt{3.csv} with RINI Local Repair. Anonymous task and source identities link the outputs and final evaluations.

All revision calls use the same developer instruction. Self-Revision receives the original and target description without an additional prior document. Retrieve-and-Revise receives the same D excerpt and explicit D--X relation supplied to RINI, and directly returns a complete proposal. RINI makes an audit call, validates the returned source/evidence quotations, and places the audit JSON in a second call requesting local replacements.

\begin{riniinlinetable}{Revision inputs and requested outputs in the supplied templates.}
\begin{tabularx}{\linewidth}{@{}P{0.22\linewidth}L L@{}}
\toprule
\rowcolor{RiniHeaderFill}
\textbf{Method} & \textbf{Evidence and instruction} & \textbf{Requested output} \\
\midrule
Self-Revision & Original text and target; no additional D & Complete revised proposal, or the original verbatim \\
Retrieve-and-Revise & D excerpt and explicit D--X relation; direct revision & Complete revised proposal, or the original verbatim \\
RINI Local Repair & Same D and target relation; source-linked audit followed by local correction & Audit JSON, then zero to four exact-span replacements \\
\bottomrule
\end{tabularx}
\end{riniinlinetable}

Retrieve-and-Revise and RINI both receive an explicit statement that the prior establishes the target contribution. The RR prompt requests a complete revised proposal. RINI requests a contribution audit followed by exact-span edit objects. The primary comparison uses the same 240 originals requiring correction under all three method-specific evaluations.

\subsection{Shared developer instruction}
\begin{promptblock}
You are revising a machine-generated scientific research proposal. Keep its research problem, technical mechanism, evaluation design, and existing attribution unless a local novelty correction requires a precise change. Never invent a result, citation, or experimental outcome.
\end{promptblock}

\subsection{Self-Revision user prompt}
\begin{promptblock}
Self-Revision control. No additional prior-work document is supplied. Audit novelty and attribution using only the original text. Do not assume that X is already known merely because it is named.

Target contribution X: {target_X}

Original proposal (revise this exact document, not a new proposal):
{original_text}

Return the complete revised proposal only, preserving its problem, technical mechanism, evaluation plan, and existing citations. If no justified change exists, return the original verbatim.
\end{promptblock}

\subsection{Retrieve-and-Revise user prompt}
\begin{promptblock}
Retrieve-and-Revise control. The same decisive prior and target relation given to RINI are supplied here; revise directly without a required atomic-claim audit. The prior establishes X, but any proposed surviving distinction Y must not be called novel without support.

Decisive prior D: {D_title} ({D_paper_id})
D excerpt as shown in Study 3: {D_context_as_shown}

Target contribution X: {target_X}

Original proposal (revise this exact document, not a new proposal):
{original_text}

Return the complete revised proposal only, preserving its problem, technical mechanism, evaluation plan, and existing citations. If no justified change exists, return the original verbatim.
\end{promptblock}

\subsection{RINI step 1: contribution-level audit}
This call does not revise the proposal. It prioritizes target-related claims and requests one to five items, each tied to a unique contiguous source quotation. A positive coverage judgment must include a verbatim quotation from the supplied D excerpt. The source and supporting strings are checked before the returned audit is used. Source containment links the returned record to the exact supplied texts.

\begin{promptblock}
RINI contribution-level audit. D establishes target X. First identify the proposal's atomic contribution claims in their original attribution scope; then judge which claim D covers, what novelty the proposal actually asserts, and what distinction Y, if any, survives the supplied D excerpt. Treat absence from this excerpt as UNCERTAIN, not proof of novelty. Do not revise the proposal yet.

X: {target_X}
D: {D_title} ({D_paper_id})
D excerpt: {D_context_as_shown}

Original proposal:
{original_text}

Return ONLY one valid JSON object with `claims` (1 to 5, prioritize claims about X). Each claim has `source_quote` (copy an exact, unique contiguous substring of the proposal, including any adjacent citation and attribution wording), `proposal_owns_claim` (boolean), `asserted_novelty_level` (integer 0..4), `D_covers_claim` (yes/no/uncertain), `D_supporting_quote` (exact substring copied verbatim from the supplied D excerpt if yes, otherwise empty), and `surviving_Y` (a concrete distinction or empty). Also include `target_X_revision_needed` (boolean) and `audit_note` (short string). Do not turn cited/adopted prior work into a proposal-owned claim. Use JSON booleans and numbers; no markdown fences or commentary.
\end{promptblock}

The direct integer novelty field in this prompt is auxiliary repair metadata. It does not reproduce the cumulative-threshold measurement used for the earlier aggregate outcome. The prompt's use of \texttt{uncertain} for D coverage also differs from \texttt{unclear} in the final evaluation schema; those fields belong to different records.

\subsection{RINI step 2: local repair}
The second call receives the audit JSON, original text, target, and D excerpt. Each proposed edit must identify an exact, unique substring of the original and give a replacement and reason. At most four edits are requested. An empty list is an allowed no-change proposal.

\begin{promptblock}
RINI local repair. Use the audited claim/evidence relations below. Withdraw or narrow proposal-owned novelty for X where D covers it; attribute the covered move to D. Keep only Y distinctions that the supplied evidence does not cover, and phrase uncertain Y as a testable possibility, not established novelty. Preserve the technical mechanism and evaluation plan. Make at most four LOCAL textual edits; do not rewrite the whole proposal and do not add a new contribution.

X: {target_X}
D: {D_title} ({D_paper_id})
D excerpt: {D_context_as_shown}

Audit JSON: {audit_json}

Original proposal:
{original_text}

Return one JSON object with `edits`, a list of 0..4 objects. Each edit has `source_quote` (exact contiguous substring of original, unique in the document), `replacement_text`, and `reason`. If the audit calls for no change, return an empty list. No text outside JSON.
\end{promptblock}

The local-repair instruction pairs attribution of known covered content with a testable formulation of unresolved distinctions. The audit is included in the second call with the original text and same prior excerpt. Exact source spans define where each replacement is applied.

\subsection{Reconstruction checks and observed dispositions}
The frozen validator accepts at most four disjoint edits. Each source quote must occur exactly once and each replacement must be nonempty. A source span is at most 1,100 characters; its replacement is at most $\max(1600,3|\mathrm{source}|)$ characters. Source spans containing the case-insensitive phrases ``evaluation plan,'' ``proposed mechanism,'' or ``methodology'' are rejected. Total replaced source characters may not exceed $\max(300,\lfloor0.35|\mathrm{original}|\rfloor)$, and the reconstructed proposal must retain at least 75\% of the original whitespace-split word count. Violation returns the unchanged source. These rules define the reconstruction gate. They are bound to the frozen prompt-and-gate code digest recorded in the execution plan.

\begin{riniinlinetable}{RINI generation dispositions across 360 source proposals.}
\begin{tabularx}{\linewidth}{@{}L r@{}}
\toprule
\rowcolor{RiniHeaderFill}
\textbf{Disposition} & \textbf{Count} \\
\midrule
Accepted local reconstruction & 291 \\
No change & 1 \\
Rejected: individual edit exceeds size cap & 4 \\
Rejected: protected section-heading overlap & 10 \\
Rejected: non-unique source or empty replacement & 23 \\
Rejected: excessive total replacement & 30 \\
Source-degenerate no-op & 1 \\
\bottomrule
\end{tabularx}
\end{riniinlinetable}

The 67 rejected reconstructions retain the source text. Each direct-revision condition records 359 generated proposals and one source-degenerate no-op. There are 359 RINI audits and one degenerate source without a substantive audit. The independent human evaluation assigns the final repair category to the returned original or reconstructed text.

\subsection{Prompt and implementation provenance}
The revision templates are reproduced from the frozen implementation at source commit \nolinkurl{0ffb836b1271ea9d2f105287abdb1508dcb83d3c}. The source package includes the revision templates and validator implementation used for the reported runs. Human annotation of the resulting revisions follows the procedure and rubric below.

\section{Human Repair Evaluation Protocol}\label{app:repair-eval-current}

\subsection{Task and evidence}
This is the final human repair-evaluation task, distinct from the pre-repair H/E diagnosis and from the internal RINI audit in Appendix~\ref{app:executed-prompts}. The 1,080 items were shuffled across methods and divided equally among five annotators, 216 items each. Each item was annotated once, without repeated annotation or adjudication. Before formal annotation, one additional colleague who did not take part in the formal round piloted 90 items, sampled as 30 from each of the three revision policies. Discussion after the trial established the rubric, including the three strict and three permissive boundaries below. Items were presented through an Azure web interface with the original proposal, revised version, target X, prior D, and its excerpt; method, generator, and original H/E identities were hidden.

The primary question is whether the revision accurately distinguishes what prior work establishes from what remains a defensible research question. An evaluator reads the entire revised proposal rather than stopping at the first corrected sentence. Core attribution can be correct in one section and contradicted elsewhere. At the same time, a minor imperfection should not by itself invalidate a repaired contribution claim.

Judgments use the supplied original, revision, target, and prior excerpt. Citation mapping, evidence uncertainty, and unreadable text have explicit labels. Evaluators assess the text under the displayed evidence and the common rubric.

\subsection{Three strict boundaries}
\paragraph{Unrepaired source attribution.}
If D establishes X, deleting ``novel'' or changing ``no method does X'' into ``whether a method can do X remains unknown'' does not resolve the prior-work error. The text must make the established provenance clear. An identifiable citation connected to the mechanism can suffice; spelling out a full title in every paragraph is not required. The verb ``propose'' alone is not proof of an ownership claim.

\paragraph{Relocating novelty to covered content.}
If the evidence establishes both a trigger X and controller Y, crediting X while claiming Y as a new mechanism still leaves a contribution problem. A change of name, notation, or granularity does not by itself create an extension. A carefully delimited experiment using Y can remain acceptable when its existing status is acknowledged.

\paragraph{Creating a new contribution error.}
A revision that removes valid provenance and newly claims an established mechanism, assigns the mechanism to the wrong source, or introduces an unsupported priority claim has regressed on contribution positioning. Deleting a redundant citation while retaining clear correct attribution is not by itself such an error. The decision concerns the final meaning of the text.

\subsection{Three permissive boundaries}
\paragraph{Peripheral empirical language.}
After the trigger is correctly attributed, a noncentral expectation such as ``this design may improve stability'' can remain an empirical hypothesis. Slightly strong wording can be recorded as a minor technical issue. A guarantee or a supposed established advantage used to support the main contribution warrants stricter scrutiny.

\paragraph{An unresolved but honestly framed Y.}
If the excerpt cannot establish whether a proposed controller is covered, use \texttt{unclear}. The proposal can still pass contribution repair if it treats the controller as a testable implementation choice rather than certified novelty. No global originality proof is required. The qualifier ``we will test'' does not excuse a false statement that D lacks a component it explicitly includes.

\paragraph{A separate technical imperfection.}
A noncentral baseline description, local terminology issue, or numerical error can be recorded separately after contribution ownership has been repaired. If the same error is used to justify a supposed new distinction, it also affects contribution repair. Severe removal of the research plan is captured in preservation and the final repair outcome.

\subsection{Evaluate X, Y, remaining problems, and preservation}
For the original and revised text, record whether substantive contribution revision is required. Attribution of X is correct, partial, incorrect, unclear, or not applicable. Partial attribution means that meaningful credit is supplied but another important passage retains a conflicting ownership claim.

Record whether Y is proposed and give its shortest faithful description. A positive \texttt{covered\_by\_d} requires evidence of substantive coverage. \texttt{not\_covered\_by\_d} requires affirmative support for the stated distinction being outside D's scope; it cannot be assigned solely because a truncated excerpt omits it. Otherwise use \texttt{unclear}. No Y uses \texttt{not\_applicable}.

Record remaining unsupported novelty and any new contribution error separately. A known error repeated from the original is a remaining error; it is not automatically a new one. \nolinkurl{other_technical_issue} describes technical issues present in the final text, not necessarily introduced by revision. It therefore cannot establish a rate of newly caused technical errors.

Assess preservation separately for the research question, method, and evaluation plan. Replacing ``we introduce X'' with ``we adopt X from D and evaluate Y'' can preserve all three. Removing the method to avoid an overclaim does not. If the original contains no usable evaluation plan, use \texttt{original\_unavailable}; this is not evidence of deletion by the reviser.

\subsection{Unchanged, truncated, and damaged outputs}
An unchanged original can remain acceptable if it had no contribution problem. It remains unrepaired if it had a known problem. A partial truncation need not make the entire item unjudgeable when the relevant contribution and method can still be evaluated; the unavailable component is labeled separately. An empty, corrupted, or substantively unreadable proposal receives \texttt{unjudgeable} and cannot become a low-novelty success.

\Needspace{20\baselineskip}
\subsection{Final outcome categories}
\detailtable[One final contribution-repair label per task.]{\textbf{Label} & \textbf{Meaning}}{
\texttt{successful\_repair} & An original substantive contribution problem is resolved, with no new substantive contribution error and useful research content retained. \\
\texttt{partial\_repair} & A meaningful contribution correction occurs, but an important contribution problem remains. \\
\texttt{no\_repair} & The central contribution problem is essentially unchanged. \\
\nolinkurl{original_already_acceptable} & The original needed no substantive contribution repair, and the revision preserves that acceptability. \\
\texttt{regressed} & Revision introduces a new substantive contribution error or causes serious loss of useful research content. \\
\texttt{unclear} & The supplied evidence cannot settle the outcome. \\
\texttt{unjudgeable} & The text does not permit a meaningful outcome judgment. \\
}

Successful contribution repair and a separately recorded technical issue can coexist. An error affecting the claimed remaining contribution or operational research plan enters the corresponding contribution or preservation judgment.

\subsection{JSONL output contract}
The result is UTF-8 JSONL: one complete JSON object per anonymous task, on one physical line, with no comments, Markdown fences, or surrounding prose. IDs are copied from the task rather than invented. All fields below are required. Evidence arrays contain verbatim source quotations; use an empty array when no quotation is available. Explanatory fields are written in English. Do not include method or treatment guesses.

Local aliases and source-file line numbers are bookkeeping fields. They can differ across exports and are not reliable cross-method join keys. The task-to-source map supplies the stable identity for pairing.

\detailtable[Allowed categorical values in the repair-evaluation JSONL.]{\textbf{Field} & \textbf{Allowed values}}{
Original contribution-revision need & \texttt{yes}, \texttt{no}, \texttt{unclear}, \texttt{unjudgeable}. \\
Revised contribution-revision need & \texttt{yes}, \texttt{no}, \texttt{unclear}, \texttt{unjudgeable}. \\
\texttt{x\_attribution} & \texttt{correct}, \texttt{partial}, \texttt{incorrect}, \texttt{unclear}, \texttt{not\_applicable}. \\
\texttt{has\_y} & \texttt{yes}, \texttt{no}. \\
\texttt{y\_vs\_d} & \texttt{covered\_by\_d}, \texttt{not\_covered\_by\_d}, \texttt{unclear}, \texttt{not\_applicable}. \\
Unsupported novelty remains & \texttt{yes}, \texttt{no}, \texttt{unclear}. \\
New contribution error & \texttt{yes}, \texttt{no}, \texttt{unclear}. \\
Other technical issue & \texttt{no}, \texttt{minor\_only}, \texttt{yes}, \texttt{unclear}, \texttt{unjudgeable}. \\
Research-question and method preservation & \texttt{preserved}, \texttt{partially\_preserved}, \texttt{lost}, \texttt{unclear}. \\
Evaluation-plan preservation & The preservation values above, plus \texttt{original\_unavailable}. \\
\texttt{repair\_outcome} & The seven final categories defined above. \\
}

The exact JSON field names, identifiers, explanations, and quotation arrays appear in the complete example and the distributed schema. If \texttt{has\_y} is no, \texttt{y\_text} is empty and \texttt{y\_vs\_d} is \texttt{not\_applicable}. Uncertainty about an honestly bounded Y is compatible with a successful repair. A new substantive contribution error is incompatible with \texttt{successful\_repair}. An original that was already acceptable cannot be counted as a newly successful repair.

\subsection{Complete constructed output example}
The following example is illustrative only and is not included in the study counts. It is displayed across lines for reading; the distributed example JSONL stores the entire object on one line. The quotation strings are from this constructed example, not from a real paper.

\begin{promptblock}
{
  "anonymous_task_id": "EXAMPLE_DO_NOT_LABEL",
  "original_file_line": 1,
  "original_alias": "O_EXAMPLE",
  "evidence_unit": "D_EXAMPLE",
  "original_contribution_revision_needed": "yes",
  "original_reason": "The original claims to introduce a trigger established by the supplied prior.",
  "revised_contribution_revision_needed": "no",
  "x_attribution": "correct",
  "has_y": "yes",
  "y_text": "A comparison of controller settings under a fixed retrieval budget.",
  "y_vs_d": "unclear",
  "unsupported_novelty_claim_remains": "no",
  "new_contribution_error": "no",
  "other_technical_issue": "minor_only",
  "technical_issue_note": "A peripheral stability expectation should remain a hypothesis.",
  "research_question_preservation": "preserved",
  "method_preservation": "preserved",
  "evaluation_plan_preservation": "preserved",
  "repair_outcome": "successful_repair",
  "original_evidence": ["We introduce a new uncertainty-triggered retrieval mechanism."],
  "revised_evidence": ["We adopt the uncertainty-triggered mechanism from D and will compare controller settings."],
  "d_evidence": ["Our method retrieves when the model uncertainty exceeds a threshold."],
  "new_issue_evidence": [],
  "preservation_note": "The trigger, controller comparison, and evaluation plan remain concrete.",
  "final_rationale": "The target receives credit and the controller is retained as a test rather than established novelty."
}
\end{promptblock}

\subsection{Formal annotation record}
The rubric was established after the 90-item pilot and discussion. Five formal annotators then labeled the 1,080 shuffled items once each. The resulting records retain the categorical outcomes and evidence fields used in the reported analyses.

\subsection{Independent final-text-only robustness annotation}
\label{app:final-only-reannotation-protocol}
This post-hoc robustness annotation is deliberately narrower than the formal original--revision evaluation above. It evaluates only the contribution state of a final proposal. Reviewers are not asked to reconstruct whether the source originally needed repair, to compare revision quality, or to identify which revision method produced the text. The fixed sample contains one selected source from each Study~3 seed and both corresponding final outputs from Retrieve-and-Revise and RINI. Pair membership is retained only for record matching and reviewer assignment; it does not reveal the expected judgment.

The primary field, \texttt{final\_contribution\_revision\_needed}, has three values. \texttt{yes} denotes a substantive remaining contribution-attribution or novelty-positioning problem that would require changing the claim, attribution, or stated distinction from prior work. \texttt{no} denotes no such substantive problem within the supplied target and decisive-prior evidence; it does not certify global novelty. \texttt{unclear} is used when the supplied evidence or final text is insufficient for a reliable yes/no decision. Each reviewer must read the complete final proposal, quote exact supporting passages, and provide a short rationale. The issued instructions are reproduced below.

The aggregation rule is applied after annotation and is not shown to reviewers. For an item reviewed by both reviewers, the source-level label is \texttt{no} only when both judgments are \texttt{no}; any \texttt{yes} judgment yields source-level \texttt{yes}; and if there is no \texttt{yes} but at least one \texttt{unclear}, the source-level label is \texttt{unclear}. An item reviewed once retains its single label. This conservative rule is used only to form the 30-source-per-method summary reported in Appendix~\ref{app:final-only-reannotation-results}.

\begin{promptblock}
ANNOTATOR INSTRUCTIONS

Your task is to assess the contribution claims in the final version of a research proposal. Read the complete final_proposal_text independently and evaluate it using target_contribution and the supplied decisive_prior evidence. Assess the final proposal as it stands. Do not assess whether an earlier version needed revision or compare the quality of revision methods.

Materials provided for each task:
- item_id: the identifier of this final proposal. Preserve it exactly in your submission.
- pair_id: a source-pair identifier used for record matching. Preserve it exactly if it is provided; otherwise the annotation system will attach it. It provides no information about the correct judgment.
- target_contribution: the target contribution whose attribution and novelty positioning should be checked.
- decisive_prior.title and decisive_prior.paper_id: the title and identifier of the prior paper used for comparison.
- decisive_prior.evidence_excerpt: the prior-paper evidence excerpt actually used in the original experiment.
- final_proposal_text: the complete final proposal, which you must read from beginning to end.

CORE QUESTION
Does this final proposal still require a substantive revision to its contribution claims?

Set final_contribution_revision_needed to exactly one of the following values:

yes
The final proposal still contains a substantive problem with contribution attribution or novelty positioning. Examples include claiming as its own new contribution a target mechanism or combination that the supplied evidence shows was already established by the prior paper; citing the prior paper while still making a materially misleading claim such as "the first," "unlike prior work," or "the prior paper lacks this mechanism"; or renaming, rephrasing, or recombining material into a combination already covered by the prior paper and presenting it as new. Use yes when resolving the problem requires changing a contribution claim, attribution, or the stated distinction from prior work.

no
Within the supplied evidence, the final proposal clearly and accurately credits the prior paper and positions its own work as a research plan, reuse, replication, application, validation, or a specific distinction supported by the evidence. It does not continue to claim an established contribution as its own new contribution. A citation, the use of a known technique, or wording such as "we propose" or "we adopt" does not determine the answer by itself; read the context to identify the actual contribution claim. Wording, formatting, or general technical-quality issues do not warrant yes unless they materially affect contribution attribution or novelty positioning. A no judgment means that you found no contribution problem requiring substantive revision within the supplied materials; it does not establish novelty across the entire research literature.

unclear
The available evidence is insufficient, the contribution positioning is materially ambiguous, or the text is damaged in a way that prevents a reliable yes or no judgment. A detail absent from the excerpt is not necessarily absent from the full prior paper or other literature. If the judgment depends on a distinction that cannot be verified from the supplied materials, identify the missing evidence and use unclear. Do not guess to force a yes or no answer.

READING AND EVIDENCE REQUIREMENTS

1. Read the complete final proposal, including relevant claims in the problem statement, method, contribution discussion, and evaluation plan. Do not assess only the contribution section or rely on keyword matching.
2. The target contribution is a focus for checking, not an assumption that the proposal is wrong. Base your judgment on the final text and the supplied evidence.
3. One citation does not automatically make all subsequent novelty claims correct. Conversely, ordinary research hypotheses, expected experimental outcomes, and clearly attributed reuse of existing techniques are not automatically contribution errors.
4. Include one to three short verbatim quotations from final_proposal_text. When relevant, include one to two short verbatim quotations from decisive_prior.evidence_excerpt. Explain how they support your judgment. Copy quotations exactly from the corresponding fields; do not present paraphrases as quotations or invent missing text. For unclear, you may leave the relevant quotation array empty if no supporting passage can be located, but explain what is missing in your reason.
5. In reason, write two to four sentences explaining the proposal's contribution claim, its relationship to the supplied prior-paper evidence, and why substantive revision is or is not needed. For no, explain why the contribution attribution and positioning are sufficiently clear.
6. Do not consult the original proposal, previous labels, method information, internal audits, or another annotator's judgments. Do not search for additional literature. Use unclear when the supplied materials are insufficient. If you encounter another final proposal derived from the same source, judge each text independently; you do not need to choose a winner or give them different labels.
7. If the interface fails to display the complete final text, report the problem and wait for the full text to be restored before submitting. Set full_text_read to true only after reading the complete displayed proposal.

FINAL SUBMISSION FORMAT

Submit a UTF-8 JSONL file with one valid JSON object per evaluated item, on one line. Do not submit a JSON array, Markdown code fences, extra tables, or explanations outside the JSONL records. Use the reviewer_id assigned by the organizer, either reviewer_01 or reviewer_02, consistently throughout your submission.

Every record must contain exactly these fields:

{
  "item_id": "Copy the task's item_id exactly",
  "pair_id": "Copy the task's pair_id exactly, or retain the value attached by the system",
  "reviewer_id": "Use your assigned reviewer_01 or reviewer_02 identifier",
  "full_text_read": true,
  "final_contribution_revision_needed": "Use exactly one value: yes, no, or unclear",
  "proposal_quotes": ["A short verbatim quotation from final_proposal_text"],
  "evidence_quotes": ["A short verbatim quotation from decisive_prior.evidence_excerpt"],
  "reason": "Two to four sentences explaining your judgment"
}

This is a field template, not a completed answer to copy literally. Replace the placeholder strings with actual identifiers, your selected label, exact quotations, and your reason. In the submitted file, write each complete object on a single line. Escape any line breaks inside strings using valid JSON syntax. proposal_quotes and evidence_quotes must be arrays of strings; they may be empty for unclear under the evidence rules above. full_text_read must be the JSON boolean true, not the string "true".

Before submitting, confirm that you have read every assigned final text in full; submitted each assigned item_id once under your reviewer_id; preserved the correct pair_id; used only yes, no, or unclear; copied quotations exactly from their corresponding fields; and omitted method names, judgments about the original proposal, previous labels, internal audits, and speculation about another annotator's decisions.
\end{promptblock}

\clearpage
\appmodule{Human Preparation Handouts}{app:issued-handouts}
The following preparation handouts retain the issued task wording, response categories, examples, and decision rules used for seed vetting, evidence curation, and qualification. Formatting and line wrapping are adjusted for the page.
\subsection{Neutral seed vetting}
\begin{promptblock}
# Neutral Seed Vetting

Judge the seed as written, without searching for prior work. Do not assess whether the idea is novel.

Fill every labelled column on every row. Allowed values:

- `scope_fit`: `borderline` / `in_scope` / `out_of_scope` - Is this a coherent ML research problem, in scope for an ICLR-level study?
- `specificity`: `needs_tightening` / `specific` / `too_broad` - Is the method direction technically specific enough to act on?
- `auditability`: `auditable` / `borderline` / `not_auditable` - Could later evidence curation audit claims made from this seed?
- `source_support_judgment`: `not_applicable` / `partial` / `strong` / `weak` - How well does the stated problem support the method direction?
- `duplicate_or_redundant`: `no` / `yes` - Is this seed a duplicate or near-duplicate of another in the packet?
- `review_confidence`: `high` / `low` / `medium` - Your confidence in this row's judgments.
- `unprimed_by_generation`: `no` / `uncertain` / `yes` - Does the seed read as neutral - NOT already leaning toward novelty, firstness, retrieval-boundary, or prior-work language? Answer no if it primes any of those.
- `review_decision`: `accept` / `reject` / `revise`

Decision rules (the intake enforces these):

- `accept` must not contradict your own labels: a seed marked `duplicate_or_redundant: yes`, `scope_fit: out_of_scope`, `auditability: not_auditable`, or `unprimed_by_generation: no` cannot be accepted.
- `revise` requires a concrete rewrite in `revised_problem` and/or `revised_method_direction`.
- `reject` requires a `rejection_reason`.
- `reviewer_notes` / `human_notes` are free text; use them for any `revise` or `reject`.

Do not edit the `source` column or any other pre-filled column.

The first data row of `neutral_seed_vetting.csv` is a completed example whose `annotation_id` starts with `EXAMPLE_DO_NOT_LABEL`. Do not edit it. It may remain in the returned file because the intake pipeline excludes it automatically from coverage, agreement, adjudication, and analysis. The separate example CSV is a read-only duplicate for quick reference.
\end{promptblock}
\subsection{Evidence curation: first issued round}
\begin{promptblock}
# Evidence Curation -- Protocol (this task only)

You are choosing the evidence for one research seed at a time. Each row is
one seed: a `problem` and a `method_direction`. The candidate papers for
that row are listed by opaque id in `CANDIDATES.md`, with their text in
`CANDIDATES.json`. You will not be told which paper is which; judge only
what is in front of you.

## What the decisive prior is

The candidate that already covers the seed's central technical move. Not
"related", not "same area": if this paper is on the desk, the seed's method
direction is no longer new. Name that move in `technical_slot` (a few
words), say why in `decisive_rationale`, and paste the sentence that caps
it into `decisive_verified_span`.

`decisive_verified_span` must be copied VERBATIM from that candidate's text
-- the same characters, no ellipsis, no paraphrase, no re-typing from memory.
Any sentence listed under the candidate in `CANDIDATES.md` qualifies as it
stands; so does any passage of its sections in `CANDIDATES.json`, which is
where to look for a sentence the list does not show. A span from anywhere
else cannot be verified and the row will be refused.

## What the hard negative is

A candidate on the SAME topic that does NOT cover the technical slot. It
should be tempting and wrong: a weak, off-topic pick makes the study's
control arm trivially easy, and a pick that arguably does cover the slot
makes it trivially hard. Paste a verbatim span into
`hard_negative_verified_span`, write the one-line
`hard_negative_covering_sentence` naming what it WOULD have to say to cover
the slot, and explain in `hard_negative_rationale` why it does not.

## The control documents

- `chosen_filler_id` -- one neutral filler from the filler list.
- `chosen_placebo_id` -- one cross-topic paper from the placebo list.
- `chosen_background_ids` -- copy EVERY id in that row's
  `background_candidate_ids`, separated by spaces. The list already holds
  exactly the number of background ids an included row needs, and the
  completed example shows the same number; if a background candidate is
  unusable, exclude the seed rather than shorten the list.

The candidates within each list are in a shuffled order that is fixed per
packet; position carries no information about how the machine ranked them.

## Disposition

- `disposition`: `included_evidence_eligible` / `excluded_not_evidence_eligible` - include only when you found BOTH a decisive prior and a hard negative you would defend.

When you exclude, leave every chosen id and span empty and put one of these
in `exclusion_reason`:

| reason | when |
| --- | --- |
| no_decisive_evidence | no candidate covers the slot |
| no_valid_hard_negative | no same-topic non-covering candidate |
| insufficient_candidate_pool | the pool is too thin to judge |
| role_collision_or_duplicate | one paper would fill two roles |
| cutoff_violation | the only decisive candidate is too recent |
| verified_span_unavailable | no readable text to quote from |
| context_budget_failure | the span cannot be shown in context |

A row that proposes `verified_span_unavailable` in `CANDIDATES.md` is a
machine suggestion, not a decision: confirm or overrule it.

## Rules

1. Work alone. Do not discuss rows with other curators or view their answers.
2. Do not use generative AI or any other assistance to choose or write answers.
3. Do not look the candidates up. Judge the text in the packet.
4. The first data row is a completed example (`EXAMPLE_DO_NOT_LABEL`); do not edit it.
5. Do not edit any column other than the answer columns named above.
6. When finished, fill `attestation.template.json` as `attestation.json` -- see `ATTESTATION_REQUIRED.md`.
\end{promptblock}
\subsection{Evidence curation: second issued round}
\begin{promptblock}
# Evidence Curation -- Protocol (this task only)

You are choosing the evidence for one research seed at a time. Each row is
one seed: a `problem` and a `method_direction`. The candidate papers for
that row are listed by opaque id in `CANDIDATES.md`, with their text in
`CANDIDATES.json`. You will not be told which paper is which; judge only
what is in front of you.

## What the decisive prior is

The candidate that already covers the seed's central technical move. Not
"related", not "same area": if this paper is on the desk, the seed's method
direction is no longer new. Name that move in `technical_slot` (a few
words), say why in `decisive_rationale`, and paste the sentence that caps
it into `decisive_verified_span`.

`decisive_verified_span` must be copied VERBATIM from that candidate's text
-- the same characters, no ellipsis, no paraphrase, no re-typing from memory.
Any sentence listed under the candidate in `CANDIDATES.md` qualifies as it
stands; so does any passage of its sections in `CANDIDATES.json`, which is
where to look for a sentence the list does not show. A span from anywhere
else cannot be verified and the row will be refused.

## What the hard negative is

A candidate on the SAME topic that does NOT cover the technical slot. It
should be tempting and wrong: a weak, off-topic pick makes the study's
control arm trivially easy, and a pick that arguably does cover the slot
makes it trivially hard. Paste a verbatim span into
`hard_negative_verified_span`, write the one-line
`hard_negative_covering_sentence` naming what it WOULD have to say to cover
the slot, and explain in `hard_negative_rationale` why it does not.

## The control documents

- `chosen_filler_id` -- one neutral filler from the filler list.
- `chosen_placebo_id` -- one cross-topic paper from the placebo list.
- `chosen_background_ids` -- copy EVERY id in that row's
  `background_candidate_ids`, separated by spaces. The list already holds
  exactly the number of background ids an included row needs, and the
  completed example shows the same number; if a background candidate is
  unusable, exclude the seed rather than shorten the list.

The candidates within each list are in a shuffled order that is fixed per
packet; position carries no information about how the machine ranked them.

## Disposition

- `disposition`: `included_evidence_eligible` / `excluded_not_evidence_eligible` - include only when you found BOTH a decisive prior and a hard negative you would defend.

When you exclude, leave every chosen id and span empty and put one of these
in `exclusion_reason`:

| reason | when |
| --- | --- |
| no_decisive_evidence | no candidate covers the slot |
| no_valid_hard_negative | no same-topic non-covering candidate |
| insufficient_candidate_pool | the pool is too thin to judge |
| role_collision_or_duplicate | one paper would fill two roles |
| cutoff_violation | the only decisive candidate is too recent |
| verified_span_unavailable | no readable text to quote from |
| context_budget_failure | the span cannot be shown in context |

A row that proposes `verified_span_unavailable` in `CANDIDATES.md` is a
machine suggestion, not a decision: confirm or overrule it.

## Rules

1. Work alone. Do not discuss rows with other curators or view their answers.
2. Do not use generative AI or any other assistance to choose or write answers.
3. Do not look the candidates up. Judge the text in the packet.
4. The first data row is a completed example (`EXAMPLE_DO_NOT_LABEL`); do not edit it.
5. Do not edit any column other than the answer columns named above.
6. When finished, fill `attestation.template.json` as `attestation.json` -- see `ATTESTATION_REQUIRED.md`.
\end{promptblock}
\subsection{Known-perturbation qualification}
\begin{promptblock}
# Known-Perturbation Qualification -- Protocol (this task only)

Each row shows two claims, **A** and **B**, that differ only in how strongly
they assert novelty. Judge which claim carries more **unsupported novelty
burden**: how strongly it asserts that its contribution is new, beyond what
the claim itself substantiates. "We are the first to do X" carries high
burden; "We adapt established techniques to do X" carries low burden. Judge
the assertion, not whether X is actually novel.

Fill `observed_direction` with exactly one of:

| If you judge | observed_direction |
| --- | --- |
| A carries more burden than B | `left_gt_right` |
| B carries more burden than A | `left_lt_right` |
| the two are substantively equal | `equal` |
| the displayed text cannot settle it | `unclear` |

Left is A, right is B, and *greater* means *more unsupported novelty burden*.

Worked example -- A: "We are the first to calibrate sim-to-real transfer."
B: "We adapt established techniques to calibrate sim-to-real transfer."
A asserts priority, B disclaims it: `left_gt_right`.

Do not force a direction: `unclear` is a real answer, and a forced guess is
worse than an abstention. Do not count words or compare lengths; the order of
A and B is random and carries no information. Some pairs are paraphrases that
carry the same burden -- `equal` is the right answer for those.

## Rules

1. Work alone; do not view anyone else's answers.
2. Do not use generative AI or any other assistance to choose or write answers.
3. The first data row is a completed example (`EXAMPLE_DO_NOT_LABEL`); do not edit it.
4. Edit only `observed_direction` and `human_notes`.
5. When finished, fill `attestation.template.json` as `attestation.json` -- see `ATTESTATION_REQUIRED.md`.
6. Do not open the study repository, its manifests, blinding maps or answer keys while this packet is open; work from the archive you were given and nothing else.

This packet qualifies you for the later rating tasks. It is short (60 rows)
and is scored against directions known by construction.
\end{promptblock}
\label{appendix:end}
\end{document}